\pdfoutput=1
\documentclass[final]{cvpr}

\usepackage{times}
\usepackage{epsfig}
\usepackage{graphicx}
\usepackage{amsmath}
\usepackage{amssymb}
\usepackage{amsfonts}
\usepackage{nicefrac}       % compact symbols for 1/2, etc.
\usepackage{microtype}      % microtypography
\usepackage[linesnumbered,ruled]{algorithm2e}

\SetCommentSty{mycommfont}
\usepackage{url}
\usepackage{enumerate}  
\usepackage{array}
\usepackage{booktabs}
\usepackage{multirow}
\usepackage{subfigure}

\usepackage[pagebackref=true,breaklinks=true,colorlinks,bookmarks=false]{hyperref}

\def\cvprPaperID{4993} % *** Enter the CVPR Paper ID here
\def\confYear{CVPR 2021}

\graphicspath{{./figs/}}

\begin{document}

%%%%%%%%% TITLE
\title{DST: Data Selection and joint Training for Learning with Noisy Labels}

% Authors must not appear in the submitted version. They should be hidden
% as long as the \iclrfinalcopy macro remains commented out below.
% Non-anonymous submissions will be rejected without review.

\author{
Yi Wei$^{1}$ \quad Xue Mei$^{1}$\thanks{Corresponding author.}  \quad Xin Liu$^{2*}$ \quad Pengxiang Xu$^{1}$ \\
$^1$College of Electrical Engineering and Control Science, Nanjing Tech University, China \quad\\
$^2$Beijing Seetatech Technology Co., Ltd \\
{\tt\small \{weiyi,mx,pengxiangxu\}@njtech.edu.cn \quad xin.liu@seetatech.com}
}

\maketitle

\begin{abstract}
% 1. deep learning and challenging
It is well known that deep learning is extremely dependented on a large amount of clean data.
Because of high annotation cost,
various methods have been devoted to annotate the data automatically. 
However, 
a lager number of sample noisy labels are generated in the datasets, which is a challenging problem.
% 2. our method (main)
In this paper, we propose a new method called DST for selecting training data accurately.
Specifically, DST fits a mixture model to the per-sample loss of the dataset label and the predicted label, 
and the mixture model is utilized to dynamically divide the training set into a correctly labeled set, 
a correctly predicted set and a wrong set.
Then, the network is trained with these set in the supervised learning.
Due to confirmation bias problem, 
we train the two networks alternately,
and each network establishes the data division to teach another network.
When optimizing network parameters, 
the correctly labeled and predicted sample labels are reweighted respectively by the probabilities from the mixture model, 
and a uniform distribution is used to generate the probabilities of the wrong samples.
% 3. exp + result
Experiments on CIFAR-10, CIFAR-100 and Clothing1M demonstrate that DST is the same or superior to the state-of-the-art methods.
% 4. Code
% we will add source code url when uploading code to github.
% Source code is available at \textcolor{magenta}{\url{...}}.
\end{abstract}

\section{Introduction}
\label{sec:introduction}
The remarkable success on training deep neural networks (DNNs) in various tasks relies on a large-scale dataset with the correctly labels.
However, labeling large amounts of data with high-quality annotations is expensive and time-consuming.
Although there are some alternative and inexpensive methods such as crowdsourcing ~\cite{yan2014learning,yu2018learning}, online queries~\cite{blum2003noise}
and labelling samples with the annotator~\cite{Tanno_2019_CVPR} that can annotate the large-scale datasets easily to alleviate this problem,
the samples with noisy labels are yielded by these alternative methods.
A recent study~\cite{Zhang_ICLR_2017} shows that a dataset with noisy labels can be overfitted by DNNs and leads to poor generalization performance of the model.

\begin{figure}[t]
    \centering
    \subfigure{
        \centering
        \includegraphics[width=0.4\linewidth]{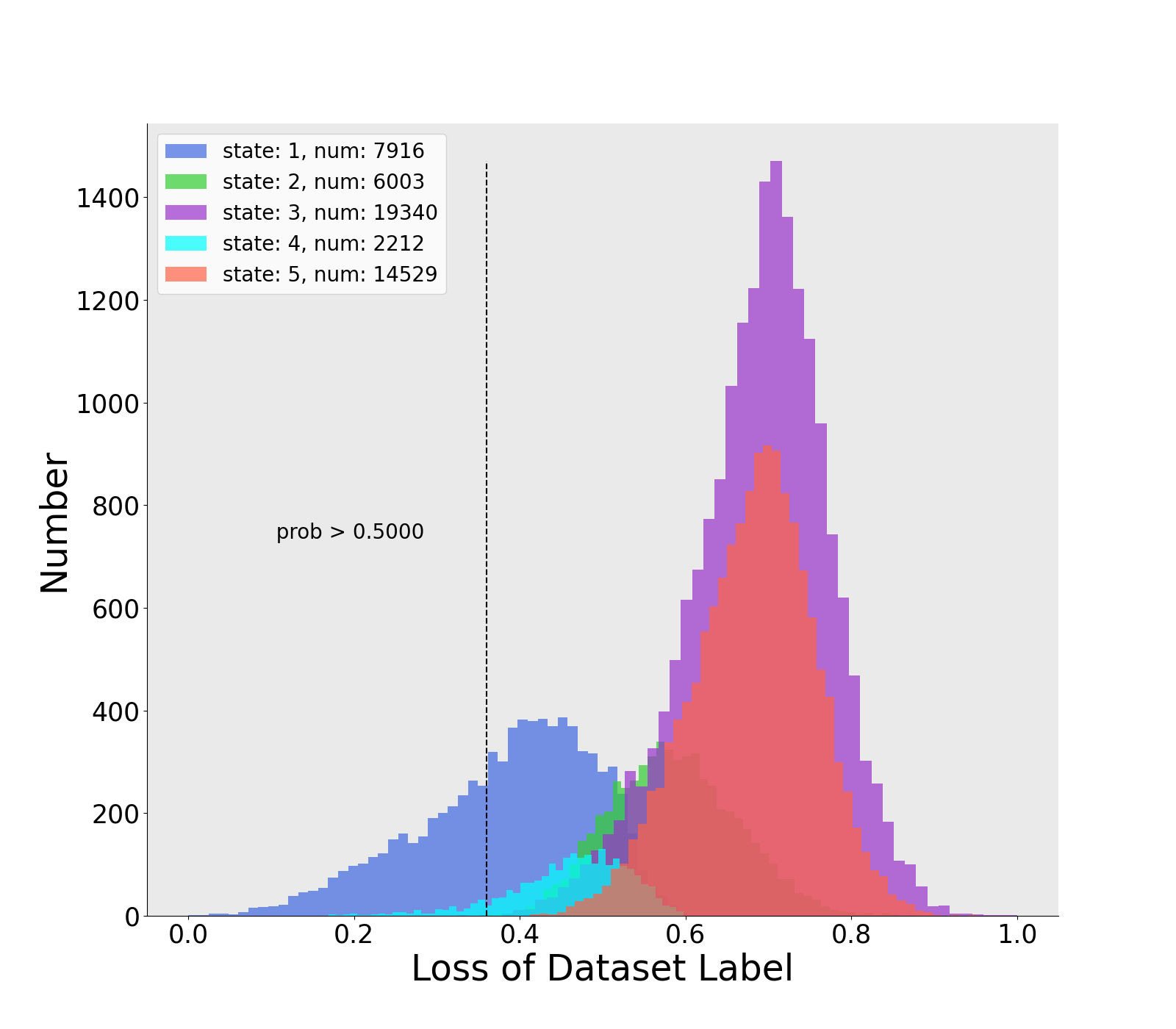}
    }
    \subfigure{
        \centering
        \includegraphics[width=0.4\linewidth]{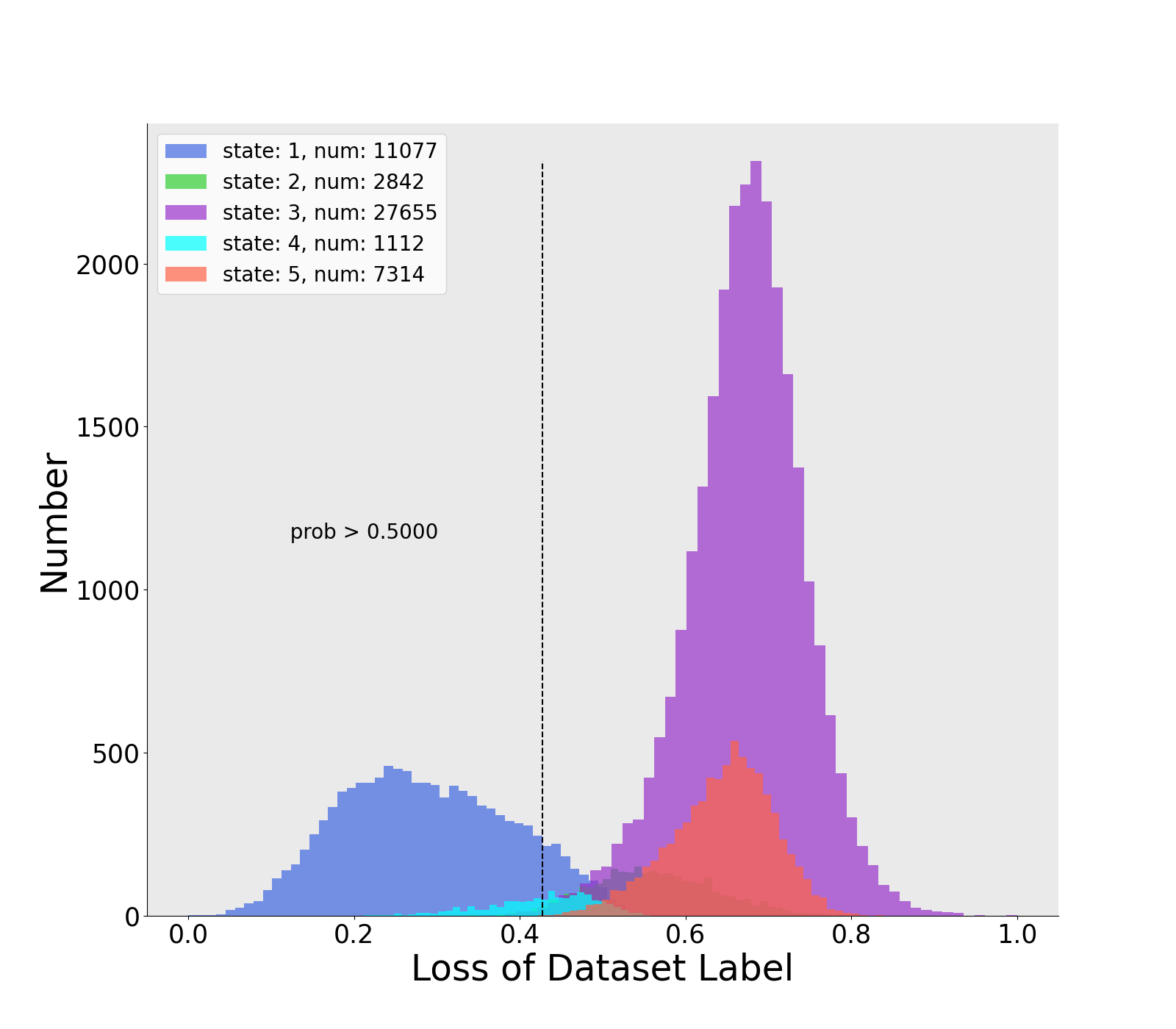}
    }
    \vspace{5pt}
    \setcounter{subfigure}{0}
    \subfigure[Warmup for 15 epochs]{
        \includegraphics[width=0.4\linewidth]{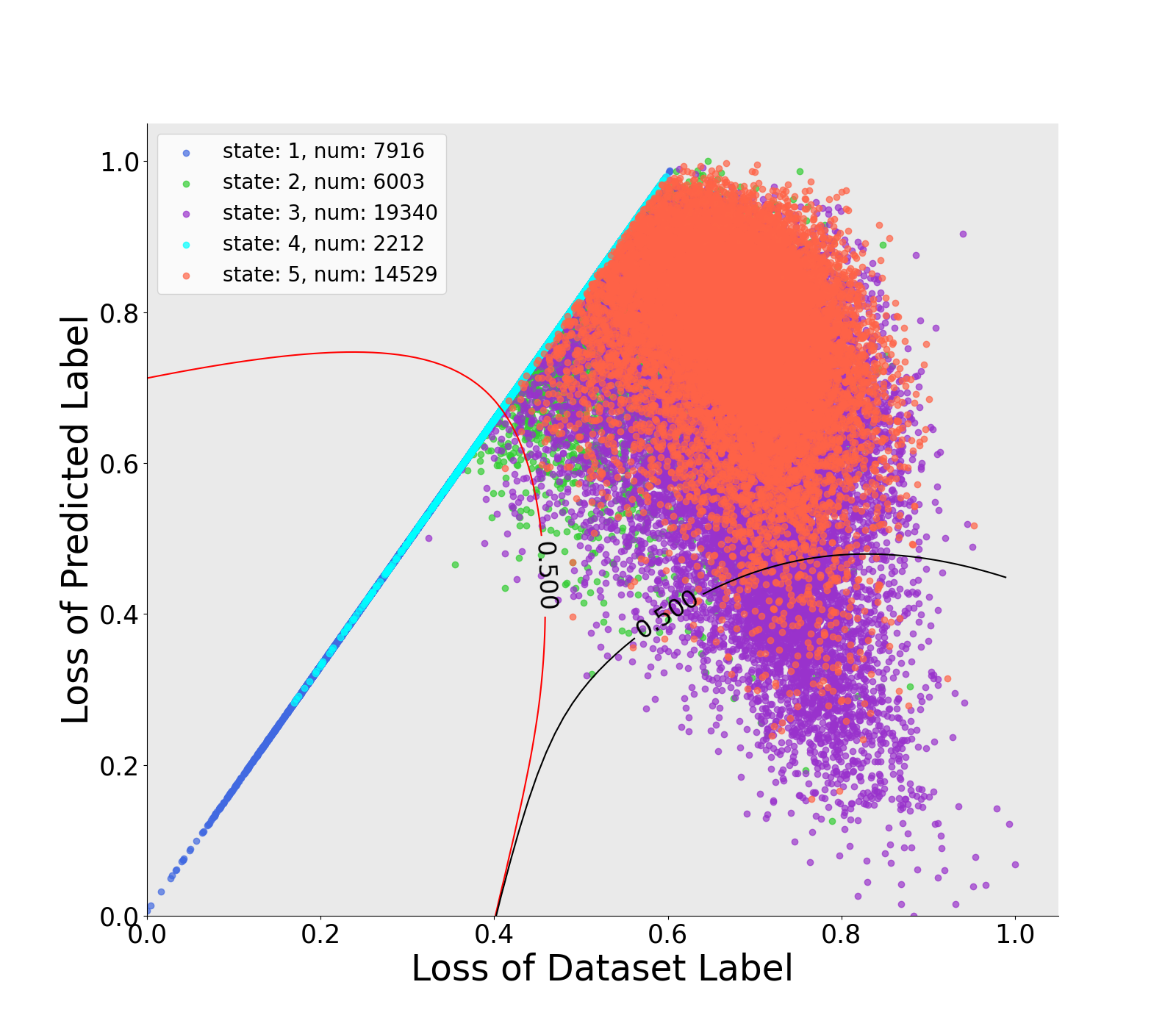}
        \centering
    }
    \subfigure[DST for 50 epochs]{
        \centering
        \includegraphics[width=0.4\linewidth]{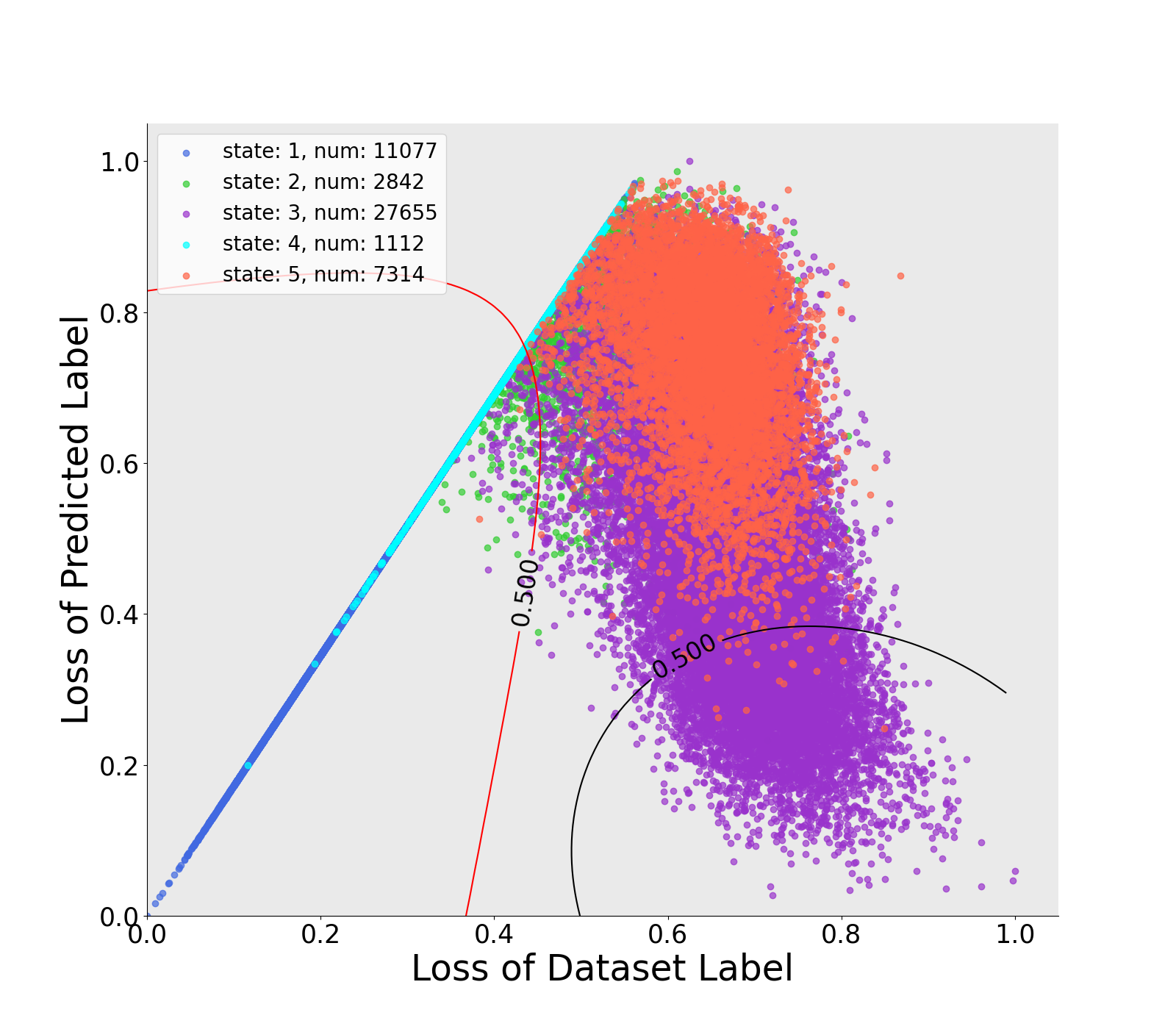}
    }
    \centering
    \caption{
        Distributions of the normalized loss on CIFAR-10 with 80\% symmetric noise.
        We use GMM to select samples for small-loss and DST.
        Top: small-loss; bottom: DST. 
        (a) 3007 samples (2816 correct samples) selected by small-loss and 7771 samples (6964 correct samples) selected by DST;
        (b) 10503 samples (9889 correct samples) selected by small-loss and 17897 samples (17054 correct samples) selected by DST.
    }
    \vspace*{-15pt}
    \label{fig:cifar_10_com}
\end{figure}

As this problem generally exists in the neural network training process and makes models get poor generalization,
there are many algorithms developed for Learning with Noisy Labels(LNL).
Some of methods attempt to estimate the latent noise transition matrix to express noisy labels and correct the loss function~\cite{Goldberger_ICLR_2017,liu2015classification,Giorgio_CVPR_2017}.
However, how to correctly construct noise transition matrix is challenging.
Some researches modify labels correctly by predictions of models for improving the model performance~\cite{Reed_2015_ICLR,Tanaka_CVPR_2018}.
Because of training labels from the DNN, the model would easily lead to overfitting under a high noise ratio.
The recent research~\cite{Arazo_ICML_2019} adopts MixUp~\cite{mixup} to address this problem.
Another approach reduces the influence of noise on the training process by selecting or weighting samples~\cite{Ren_ICML_2018}.
Many methods select clean samples with small loss~\cite{Arazo_ICML_2019,Jiang_ICML_2018}.
Co-teaching~\cite{co-teaching}, Co-teaching$+$~\cite{Yu_ICML_2019} and JoCoR~\cite{Wei_2020_CVPR} use two networks to select small-loss samples to train each other.

Despite small-loss is a good method to choose correct samples from the noise samples,
the samples which is predicted correctly by the models are ignored in training process. 
In this work,
we propose DST (Data Selection and joint Training), which can leverage correctly predicted samples and avoid overfitting new labels chosen by the model under a high level of the noise ratio.
Compared with other methods using small loss~\cite{Arazo_ICML_2019,co-teaching,Yu_ICML_2019},
we propose another method based on two kinds of the loss on each sample to distinguish samples with correctly labels for training networks.
We provide experiments to demonstrate the feasibility of our approach, which is superior to many related approaches.
Our main contributions are as follows:
\vspace{-\topsep}
\begin{itemize}
    \setlength\itemsep{0pt}
    \item
          We propose two kinds of the sample loss (1. loss of the label from the dataset; 2. loss of the label predicted by model.), % 修改
          which can be used to distinguish correctly labeled and predicted samples.
          We fit a Gaussian Mixture Model (GMM) dynamically on dataset loss distribution to divide the dataset into correctly labeled samples,
          correctly predicted samples and wrong samples with wrong labels and predictions.
    \item
          We train two networks to generate losses of samples.
          For each network, we use GMM to get correct samples,
          which is then used to train another network.
          This can filter different types of error and avoid confirmation bias in self-training~\cite{li2020dividemix}.
\end{itemize}
\vspace{-\topsep}

\section{Related Work}
\label{sec:literature}

\begin{figure*}[t]
    \begin{center}
        \includegraphics[width=0.95\textwidth]{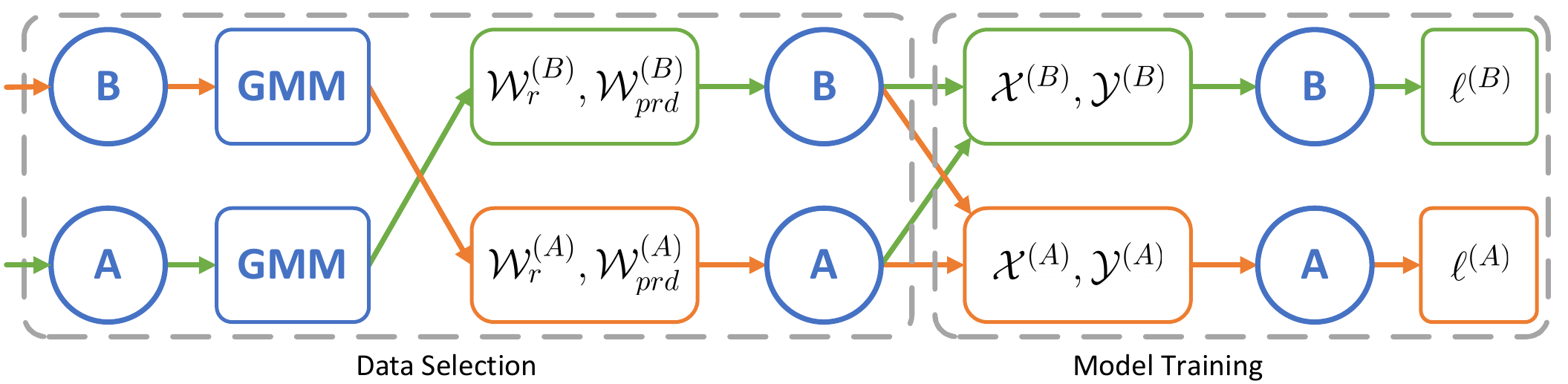}
    \end{center}
    \caption
    {
        DST uses two networks (A\&B) to teach each other.
        At each epoch, firstly,
        network A divides the data by a GMM (Data Selection), and network B uses new samples to optimize its parameters (Model training).
        Then, we training network A with the data generated by network B.
    }
\vspace*{-15pt}
\label{fig:framework}
\end{figure*}

\begin{algorithm}[t]
    \caption{\small DST} 
    \label{alg:algorithm1}
	\small
    \textbf{Input:} $\theta^{(1)}$ and $\theta^{(2)}$, 
    training dataset $(\mathcal{X},\mathcal{Y})$, 
    GMM initialization parameters $\gamma$, 
    probability threshold  $\tau_k$,
    sharpening temperature $T$, 
    regularization weight $\lambda_r$, 
    $\mathrm{Beta}$ distribution parameter $\alpha$ for $\mathrm{MixUp}$. \\

	$\theta^{(1)},\theta^{(2)}=\mathrm{Warmup}(\mathcal{X},\mathcal{Y},\theta^{(1)},\theta^{(2)})$\\
	\While{$\mathrm{epoch}<\mathrm{MaxEpoch}$}    
	{	
        $\mathcal{L}^{(1)}_r, \mathcal{L}^{(1)}_{prd}=\mathrm{CE}(\mathcal{X},\mathcal{Y},\theta^{(1)})$ \\
        $\mathcal{L}^{(2)}_r, \mathcal{L}^{(2)}_{prd}=\mathrm{CE}(\mathcal{X},\mathcal{Y},\theta^{(2)})$ \\
        $\mathcal{W}^{(2)}_r, \mathcal{W}^{(2)}_{prd}=\mathrm{GMM}(\mathcal{L}^{(1)}_r, \mathcal{L}^{(1)}_{prd}, \gamma)$ \\
        $\mathcal{W}^{(1)}_r, \mathcal{W}^{(1)}_{prd}=\mathrm{GMM}(\mathcal{L}^{(2)}_r, \mathcal{L}^{(2)}_{prd}, \gamma)$ \\
        \For {$n=1,2$}    
        {
            \For {$\mathrm{iter}=1$ \KwTo $\mathrm{num\_iters}$}
            {
                From $(\mathcal{X},\mathcal{Y},\mathcal{W}^{(n)}_r,\mathcal{W}^{(n)}_{prd})$, Draw a mini-batch $\mathcal{B}=\{(x_b,y_b,w_b^r,w_b^{prd});b \in(1,...,B)\}$ \\
                \For {$b=1$ \KwTo $B$}
                {
                    $p_b=\frac{1}{2}\big(p(x_b;\theta^{(1)}) + p(x_b;\theta^{(2)})\big)$ \\
                    \uIf{$w_b^r\ge\tau_r$}
                    {
                        $\tilde{y}_b=w_b^r y_b+(1-w_b^r)p_b$ \\
                    }\uElseIf{$w_b^{prd}\ge\tau_{prd}$}
                    {
                        $\tilde{y}_b=(1-w_b^{prd}) y_b+w_b^{prd}p_b$ \\
                    }\Else
                    {
                        $w_b^u\sim\mathcal{U}(0,1)$  \\
                        $\tilde{y}_b=(1-w_b^u) y_b+w_b^up_b$ \\
                    }
                    $\hat{y}_b=\mathrm{Sharpen}(\tilde{y}_b,T)$ 	\\
                }
                $\hat{\mathcal{B}}=\{(\hat{x}_b,\hat{y}_b);b\in(1,...,B)\}$ \\
                $\mathcal{L}_{\hat{\mathcal{B}}}=\mathrm{MixUp}(\hat{\mathcal{B}},\alpha)$ \\
                $\mathcal{L}=\mathcal{L}_{\hat{\mathcal{B}}}+\lambda_r \mathcal{L}_\mathrm{reg}$ \\
                $\theta^{(n)}=\mathrm{SGD}(\mathcal{L},\theta^{(n)})$ \\
            }
        }
    }
\end{algorithm}

In this section, we briefly review existing approaches for LNL.

\subsection{Correction model}
Correction model is to seek the noisy labels to correct the loss function. 
One way is to relabel the noisy samples to correct the loss.
Some of methods need a set of clean samples to model noisy samples with knowledge graph ~\cite{Li_ICCV_17},
directed graphical models~\cite{Tong_CVPR_2015},
conditional random field~\cite{Vahdat_NIPS_2017} and neural networks~\cite{Andreas_CVPR_2017,Lee_CVPR_2018}.
To address the problem of the clean set, 
Some of methods attemp to relabel samples by network predictions in the process of iteration~\cite{Tanaka_CVPR_2018} and ~\cite{Yi_2019_CVPR}.
Another way is named loss correction methods, 
which can modify the loss function during training to make models more robust.
Bootstrap ~\cite{Reed_2015_ICLR} modifies the loss by comparing the raw labels with model predictions.
Ma \etal~\cite{Ma_ICML_2018} use the dimensionality of feature subspaces to improve the Bootstrap.
Backward and Forward~\cite{Giorgio_CVPR_2017} estimate two noise transition matrix,
and Hendrycks \etal~\cite{Hendrycks_NIPS_2018} use a clean set to improve these matrix for loss correction.

\subsection{Division model}
% 1. rewight
% 2. small-loss selection
Division Model is to reweight training samples or divide them into a clean set and a noisy set to update network parameters~\cite{Abstention}.
MentorNet~\cite{Jiang_ICML_2018} trains a mentor network to reweigh the samples which are given to train a student network.
Ren \etal~\cite{Ren_ICML_2018} use a meta-learning algorithm to reweight samples.
Small-loss selection is a common method to extract the clean samples which have smaller loss than noisy samples.
Co-teaching~\cite{co-teaching} trains two networks to obtain the small-loss samples and learn from each other.
Co-teaching$+$~\cite{Yi_2019_CVPR} introduces the disagreement data to improve co-teaching. 
Shen and Sanghavi~\cite{Shen_ICML_2019} use one network to select small-loss samples and provide clean samples for another network training.
Arazo \etal~\cite{Arazo_ICML_2019} reweight the samples with the small loss by fitting a mixture model. 

\subsection{Other methods}
Robust loss funcions are a simple and generic solution for LNL.
Ghosh and Kumar~\cite{Ghosh_AAAI_17} have proved that some loss functions (e.g., Mean Absolute Error) can be more robust to noisy labels than commonly used loss functions.
Wang \etal~\cite{2020Symmetric} propose the Symmetric cross entropy Learning (SL) boosting CE loss symmetrically with Reverse Cross Entropy (RCE).
Although these loss function methods can improve the robustness of the network, 
for high noise ratios, they still exist some limitations.
Semi-supervised learning methods are used in this domain recently and achieve a good performance.
Some reseaches show the possibility of semi-supervised learning in LNL~\cite{WACV,Recycling}.
DivideMix~\cite{li2020dividemix} propose small-loss methods to divide the training data into the clean labeled samples and unlabeled samples,
and use both the labeled and unlabeled data to train the models in a semi-supervised learning~\cite{mixmatch}.

Different from the aforementioned methods, 
our method divides the dateset into a correct labeled set, a correct predicted labeled set and a wrong set for supervised learning.
As in Figure~\ref{fig:cifar_10_com}, compared to small-loss method which can only distinguish clean samples, 
we model the per-sample loss of the raw label and the predicted label with a mixture model to obtain more useful samples.
Compared to DivideMix~\cite{li2020dividemix},
our method can train the network under a supervised learning with more correct labels to make the network more robust.

\section{Method}
\label{sec:method}

\begin{figure*}[!t]
    % 是否用50%
    \centering
    \subfigure{
            \centering
            \includegraphics[width=4cm,height=3.5cm]{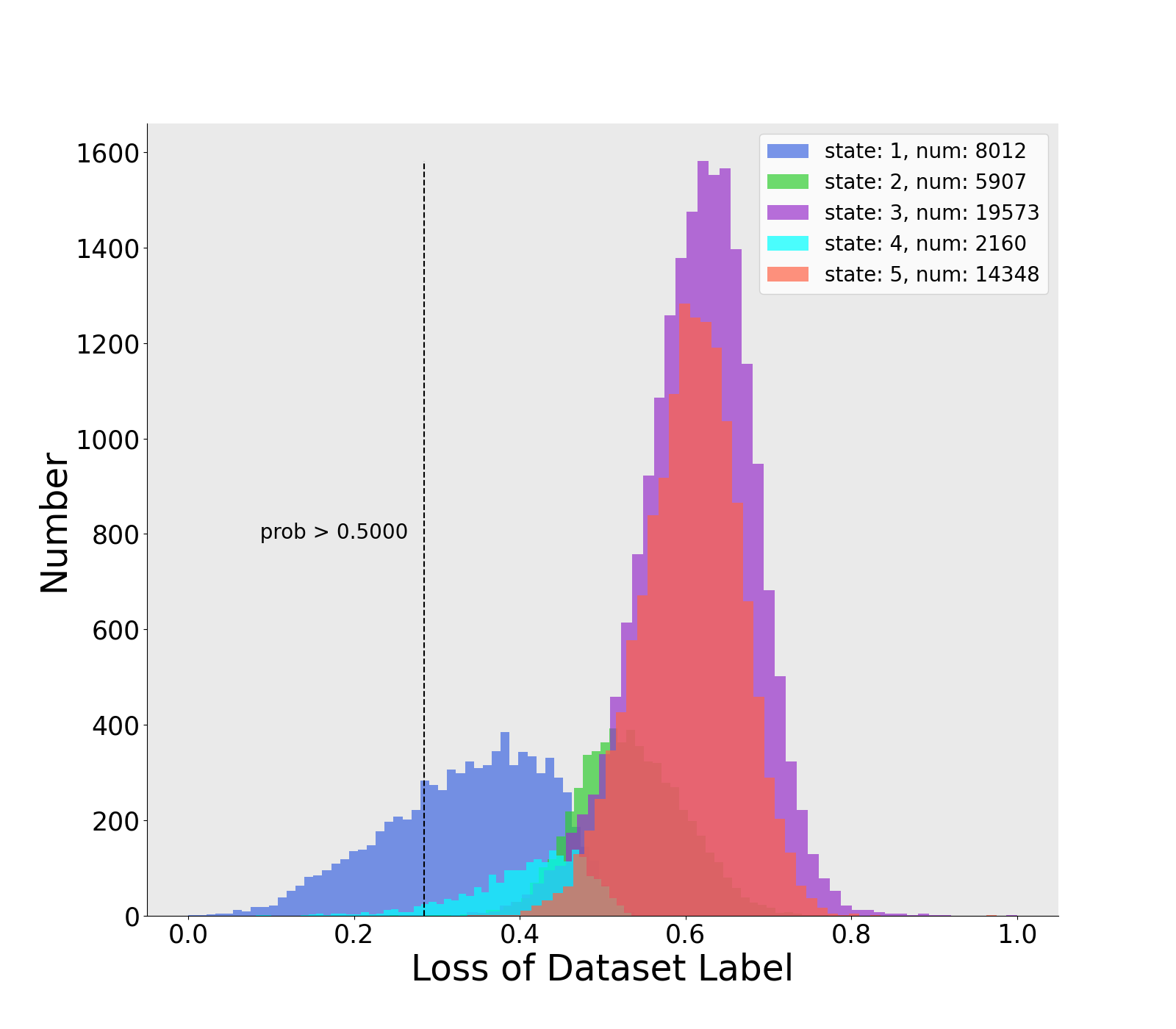}
    }
    \subfigure{
            \centering
            \includegraphics[width=4cm,height=3.5cm]{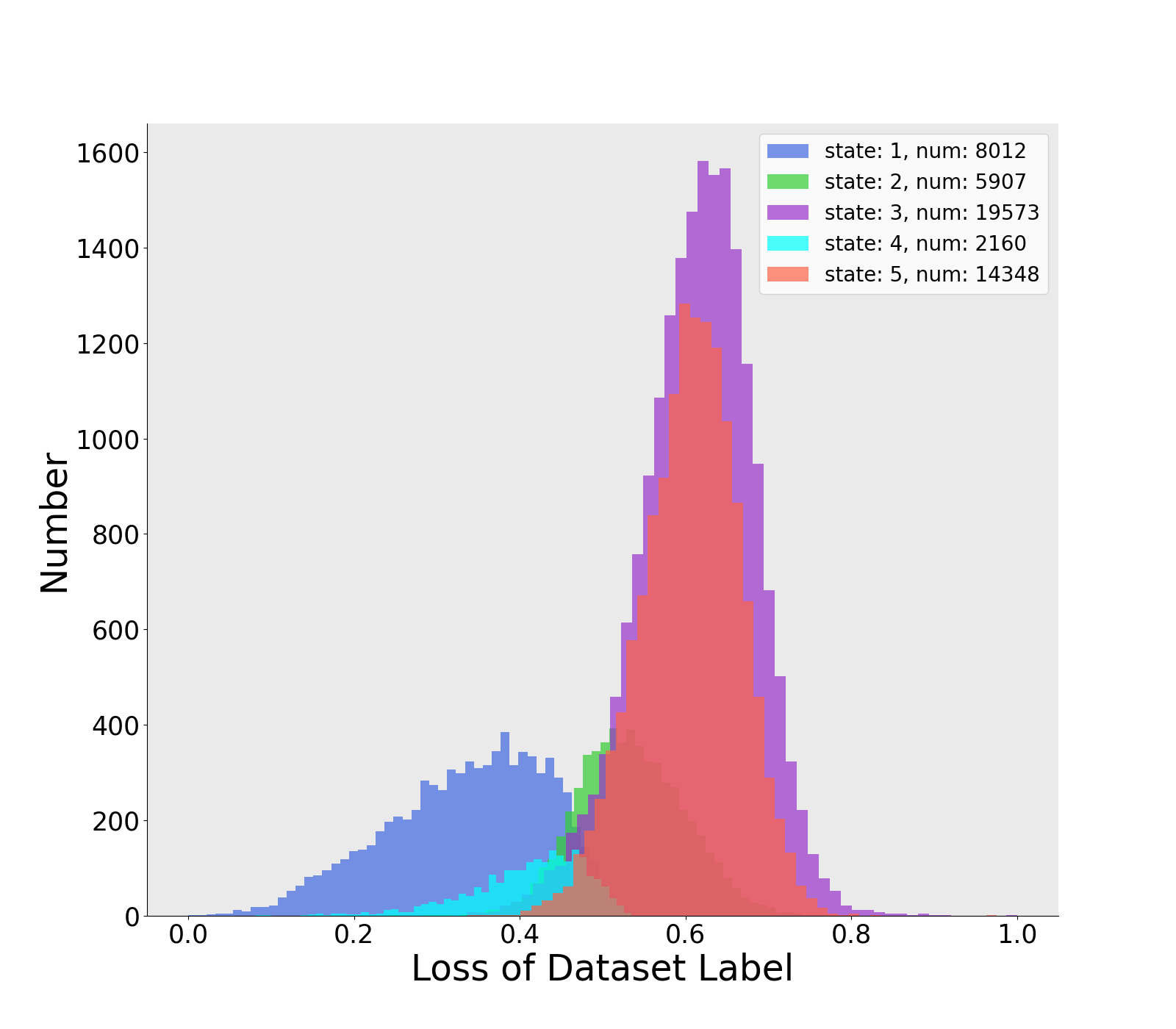}
    }
    \subfigure{
            \centering
            \includegraphics[width=4cm,height=3.5cm]{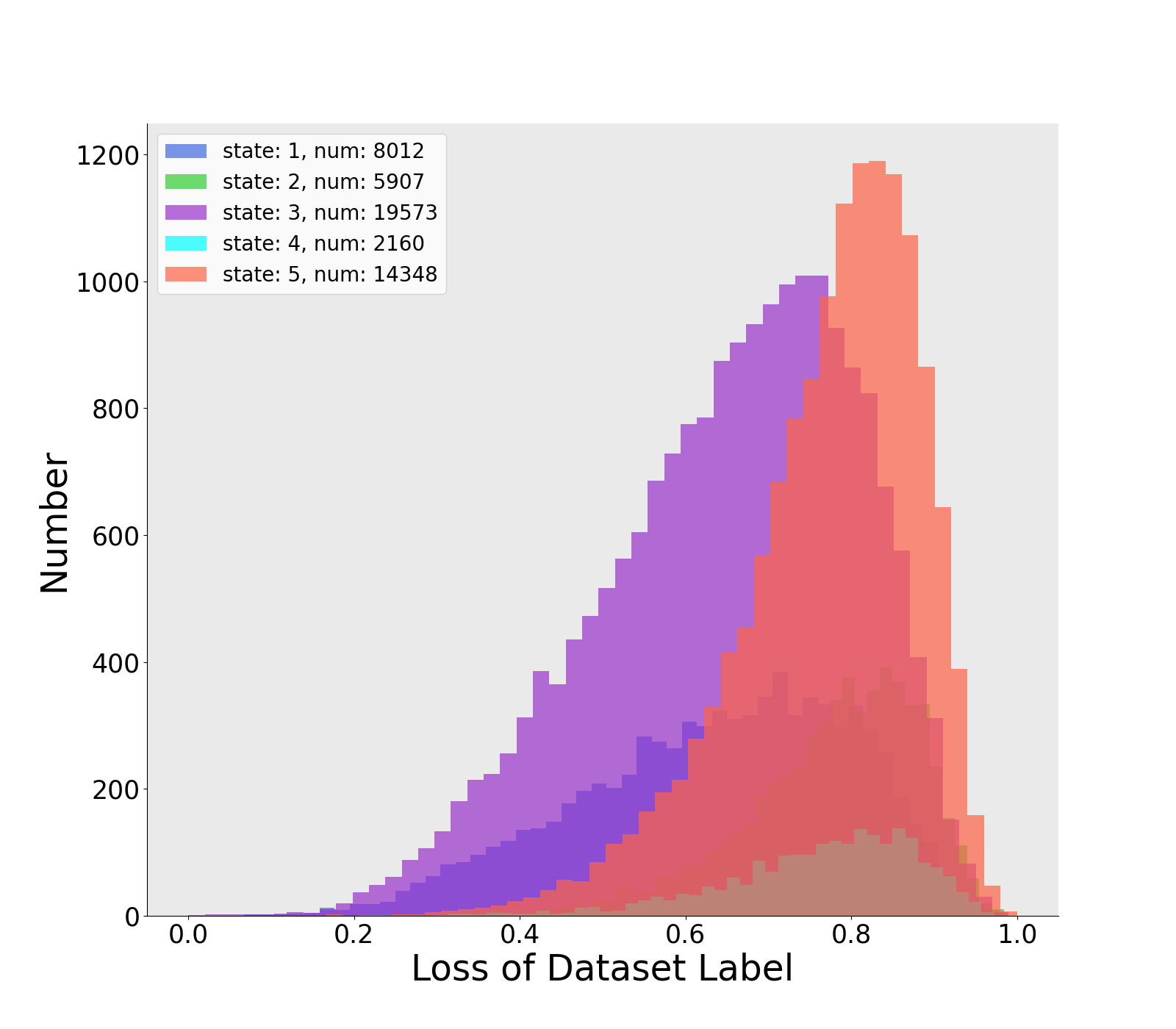}
    }
    \subfigure{
            \centering
            \includegraphics[width=4cm,height=3.5cm]{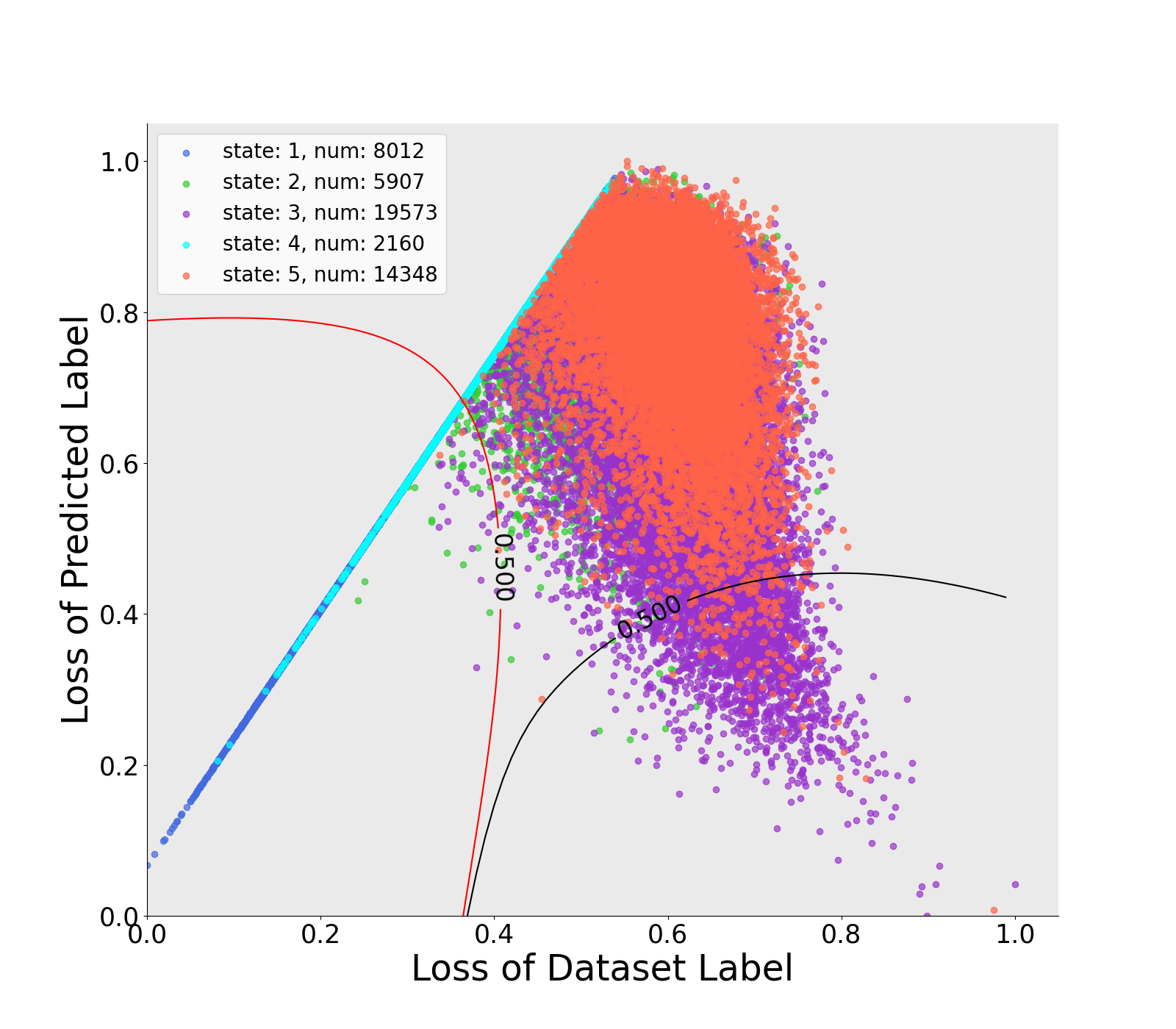}
    }

    \vspace{-10pt}
    \setcounter{subfigure}{0}
    \subfigure[small-loss]{
            \centering
            \includegraphics[width=4cm,height=3.5cm]{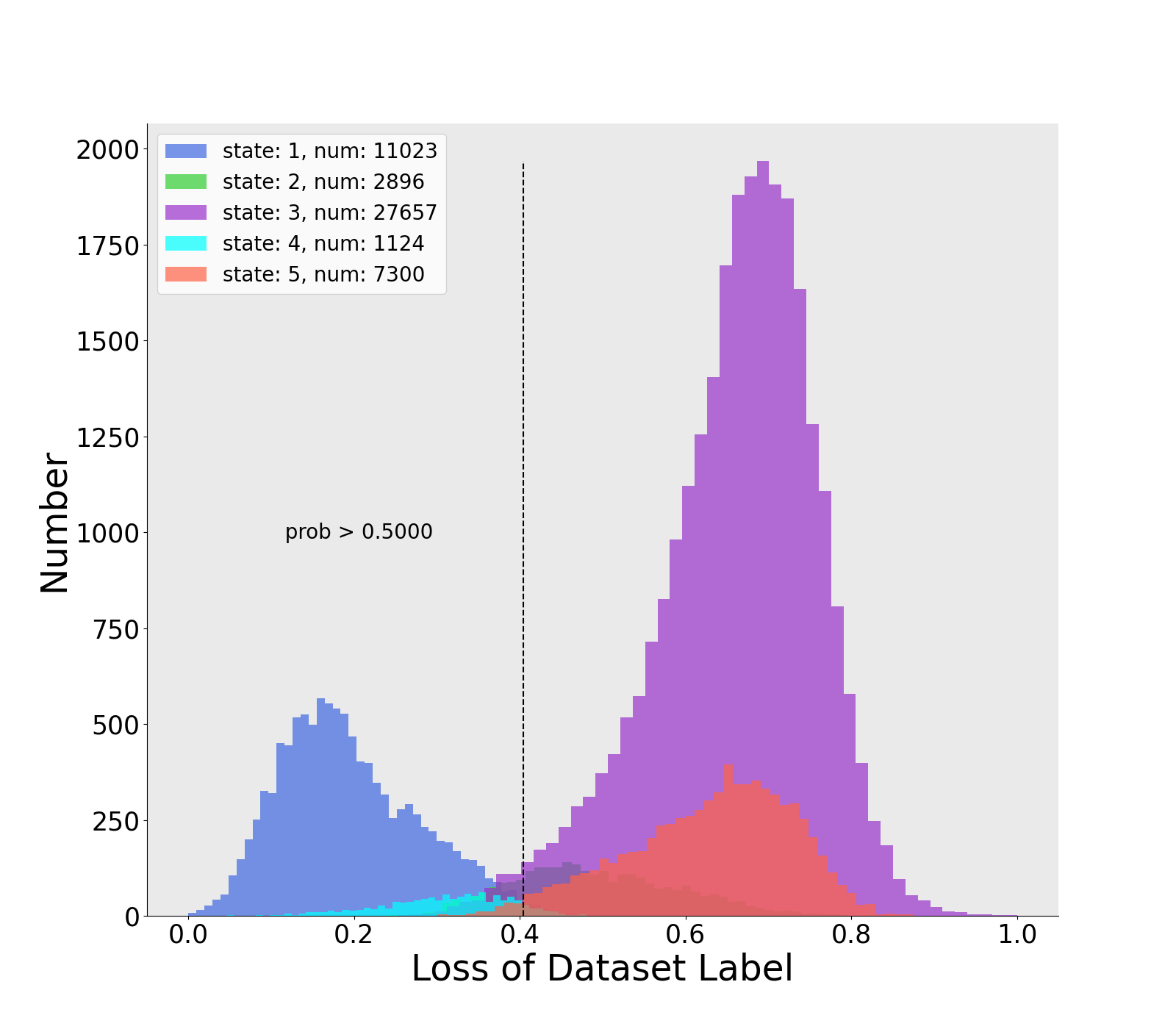}
    }
    \subfigure[loss of raw labels]{
            \centering
            \includegraphics[width=4cm,height=3.5cm]{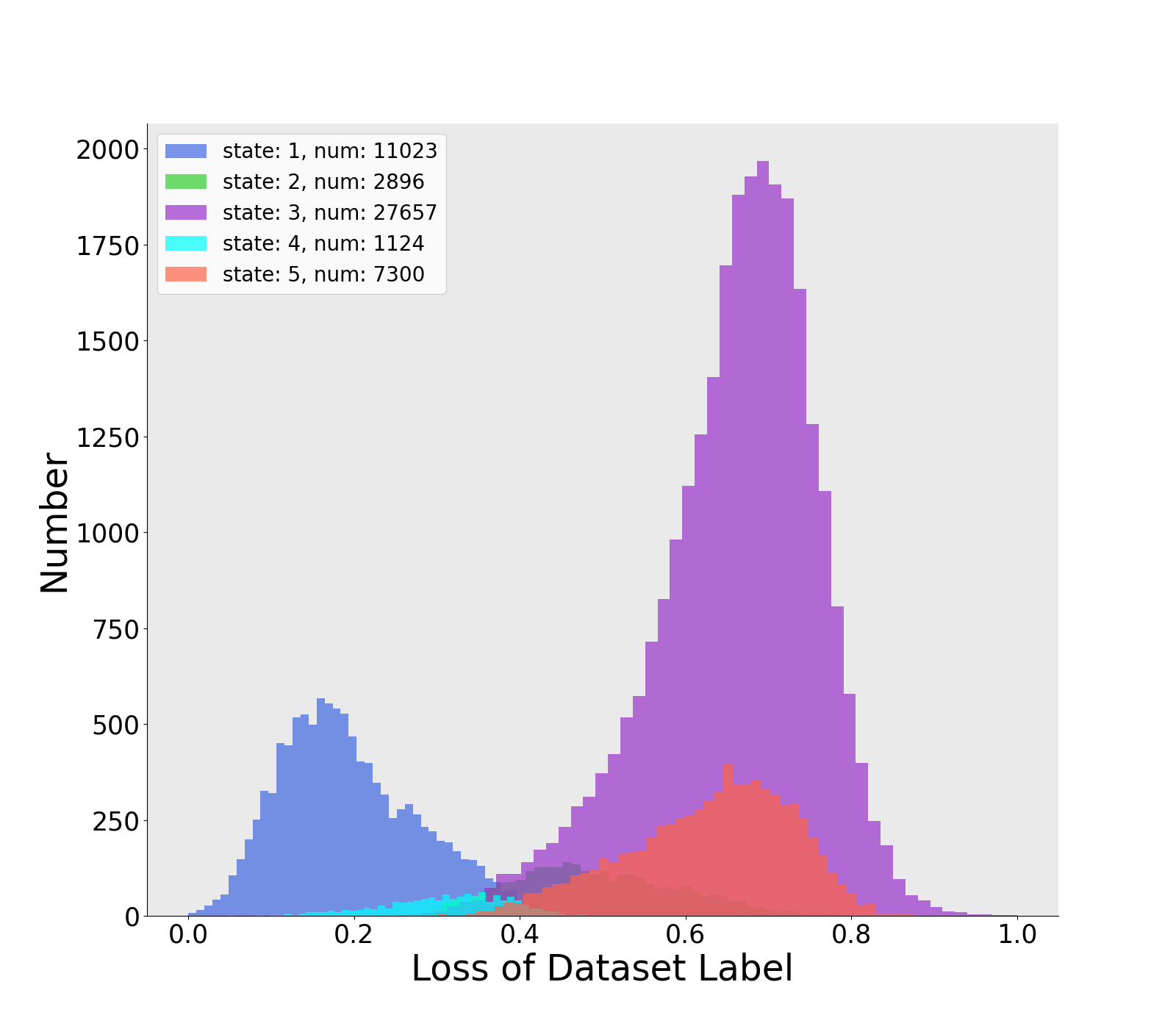}
    }
    \subfigure[loss of predicted label]{
            \centering
            \includegraphics[width=4cm,height=3.5cm]{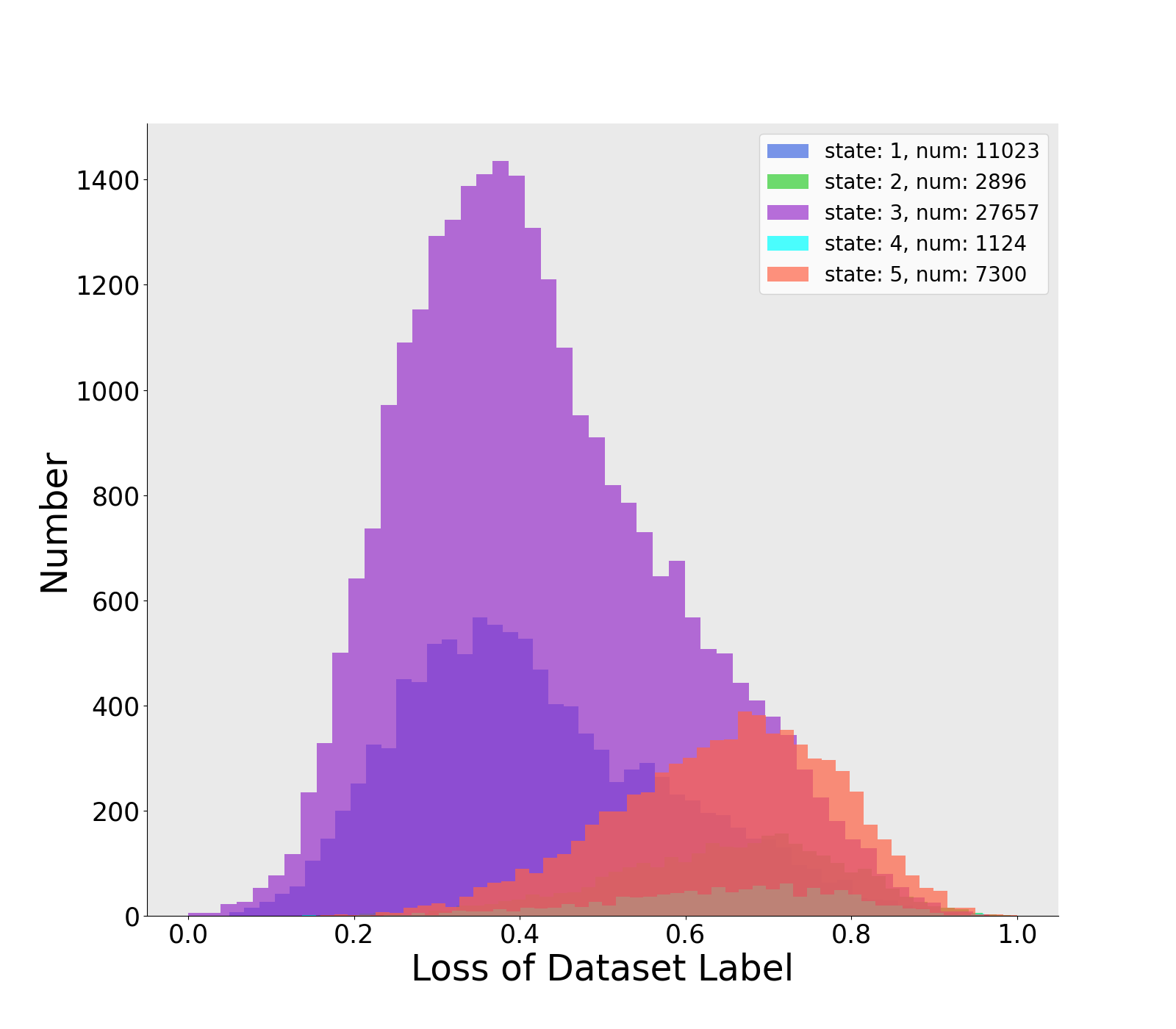}
    }
    \subfigure[DST]{
            \centering
            \includegraphics[width=4cm,height=3.5cm]{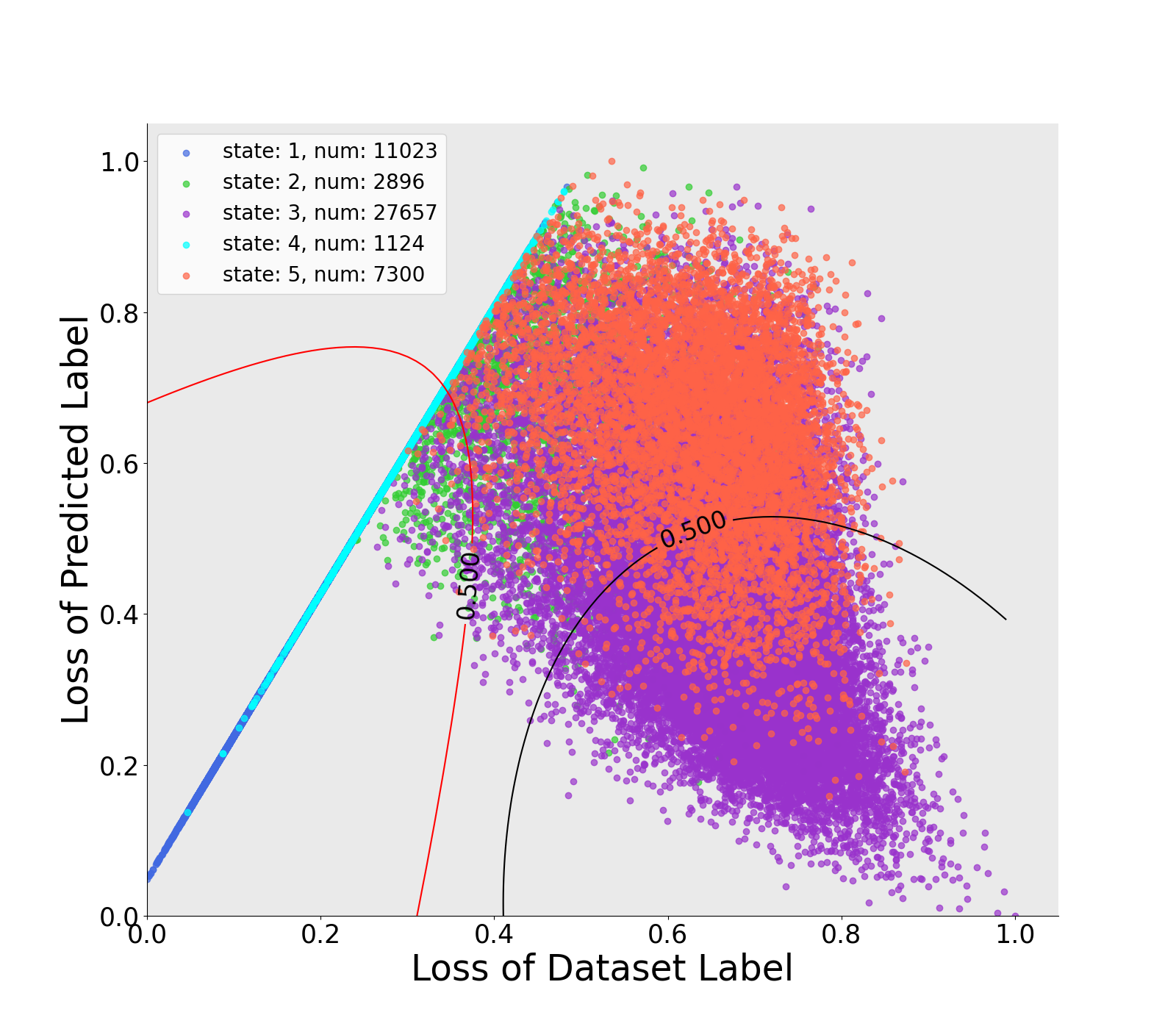}
    }
    \centering
    \caption{
        Distributions of the normalized loss on CIFAR-10 with 50\% noise ratio. 
        Top: 15 epoch; bottom: 50 epoch.
        (a) Distribution of the small-loss method;
        (b) Distribution of ;
        (c) Distribution of;
        (d) Distribution of;
    }
    \vspace*{-15pt}
    \label{fig:dis_prove}
\end{figure*}

Our approach attempts to train two deep networks simultaneously to avoid overfitting caused by self-training~\cite{li2020dividemix}.
As in Figure~\ref{fig:framework},
one network can divide the dataset into the correct data and the wrong data to teach the other network.
We use these data to supervise network learning.
DST is shown in Algorithm~\ref{alg:algorithm1}.

\subsection{Date selection}
\label{med:data_selection}

Deep networks can fit clean samples fast~\cite{Arpit_ICML_2017} and cause low loss for clean samples~\cite{Chen_ICML_2019}.
Recent researches~\cite{Arazo_ICML_2019} find the probability of clean samples to train the network by small loss.
However,
during training process,
network can produce a large number of correctly predicted samples with wrong dataset labels,
which can be used for training. 
We aim to find these samples and the samples with correct labels.

For the image classification, the dataset with $N$ samples is given as
{\small $\mathcal{D}=\{(x_i,y_i)\}_{i=1}^N$}, 
where $x_i$ is a image with $C$ classes and {\small $y_i \in \{0,1\}^C$} is the one-hot label of $x_i$.
$p_\theta$ is denoted as a model,
in which the parameters theta can be fit by optimizing the cross-entropy loss function $\ell(\theta)$:
\begin{equation}
    \ell(\theta)=\sum_{i=1}^N \ell_i^{nis}(\theta)=-\sum_{i=1}^N\sum_{c=1}^C  y_i^c \log({p_{\theta}^c}(x_i))
\end{equation}
where $p_\theta^c(x_i)$ is the softmax probabilities of the model output and $\ell_i^{nis}(\theta)$ is the loss from noisy label.
In addition to $\ell_i^{nis}(\theta)$, we also consider another kind of loss from the labels predicted by the model.
We denote this loss as $\ell_i^{prd}(\theta)$:
\begin{equation}
    \ell_i^{prd}(\theta)=-\sum_{c=1}^C  \overline{y}_i^c \log({p_{\theta}^c}(x_i))
\end{equation}
where $\overline{y}_i$ is the one-hot label generated from the model prediction of $x_i$.

For noisy dataset, let $y$, $y_{real}$ and $\overline{y}$ denote three types of sample labels,
where $y$ is a label from the noisy dataset,
$y_{real}$ is a real label and $\overline{y}$ is a label of the model prediction.
Therefore, each sample must be in one of the following five states:
\begin{enumerate}[(i)]
    \setlength\itemsep{0.005em}
    \item $y=y_{real}$, $\overline{y}=y_{real}$, $y=\overline{y}$;
    \item $y=y_{real}$, $\overline{y} \neq y_{real}$, $y \neq \overline{y}$;
    \item $y \neq y_{real}$, $\overline{y}=y_{real}$, $y \neq \overline{y}$;
    \item $y \neq y_{real}$, $\overline{y} \neq y_{real}$, $y=\overline{y}$;
    \item $y \neq y_{real}$, $\overline{y} \neq y_{real}$, $y \neq \overline{y}$.
\end{enumerate}
We suppose that small-loss instances are indeed clean and reliable,
in other words,
DNNs can learn the clean samples and ignore the noisy samples with the prediction close to an average probability (\ie $p=1/C$).
DNN is more likely to predict a samples which is closer in space to the samples learned by DNN. 
% TODO: 具体描述
If the noise sample has the close spatial distance to the correct sample learned by DNN and is predicted correctly, 
this noise sample may have a small loss of the predicted label and a large loss of the raw label (noise label). 
On the contrary, 
the noise sample may have the similar losses between the predicted label and the raw label.  
% % TODO: 句式修改
% for example,
% DNN learning a sample with $x=[0]$ tends to correctly predict a sample with $x=[1]$ rather than $x=[101]$,
% unless $x=[100]$ with the same label as $x=[0]$ has been learned by DNN.
Therefore,
we assume that the $\ell_i^{prd}$ of the correct predicted samples which are close to the learned samples is lower than the other unlearned samples.

\begin{figure*}[!t]
    \centering
    \subfigure{
            \centering
            \includegraphics[width=4cm,height=3.5cm]{cr1/Lab-info/epoch_15/net1_losses.png}
    }
    \subfigure{
            \centering
            \includegraphics[width=4cm,height=3.5cm]{cr1/Lab-info/epoch_50/net1_losses.png}
    }
    \subfigure{
            \centering
            \includegraphics[width=4cm,height=3.5cm]{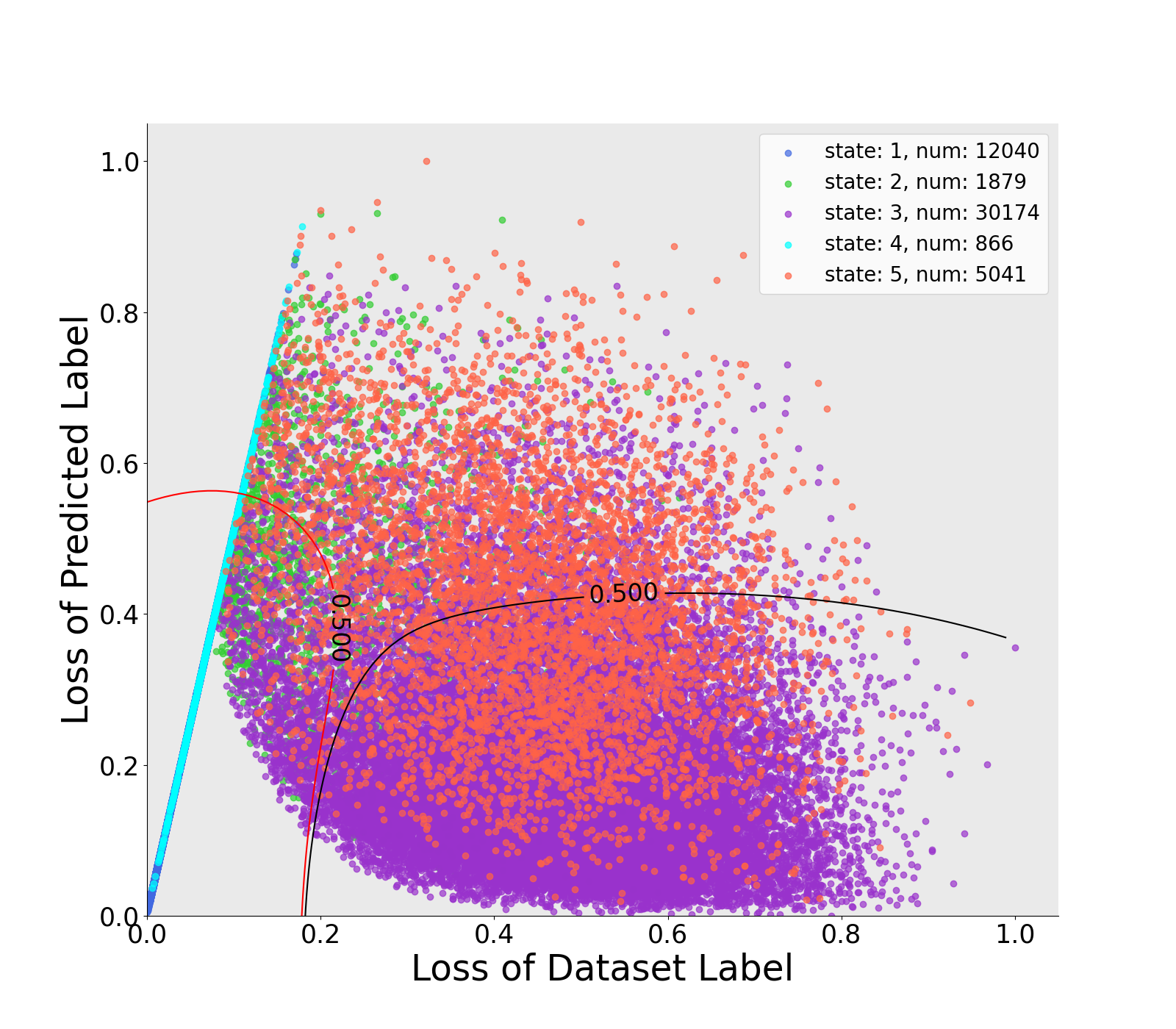}
    }
    \subfigure{
            \centering
            \includegraphics[width=4cm,height=3.5cm]{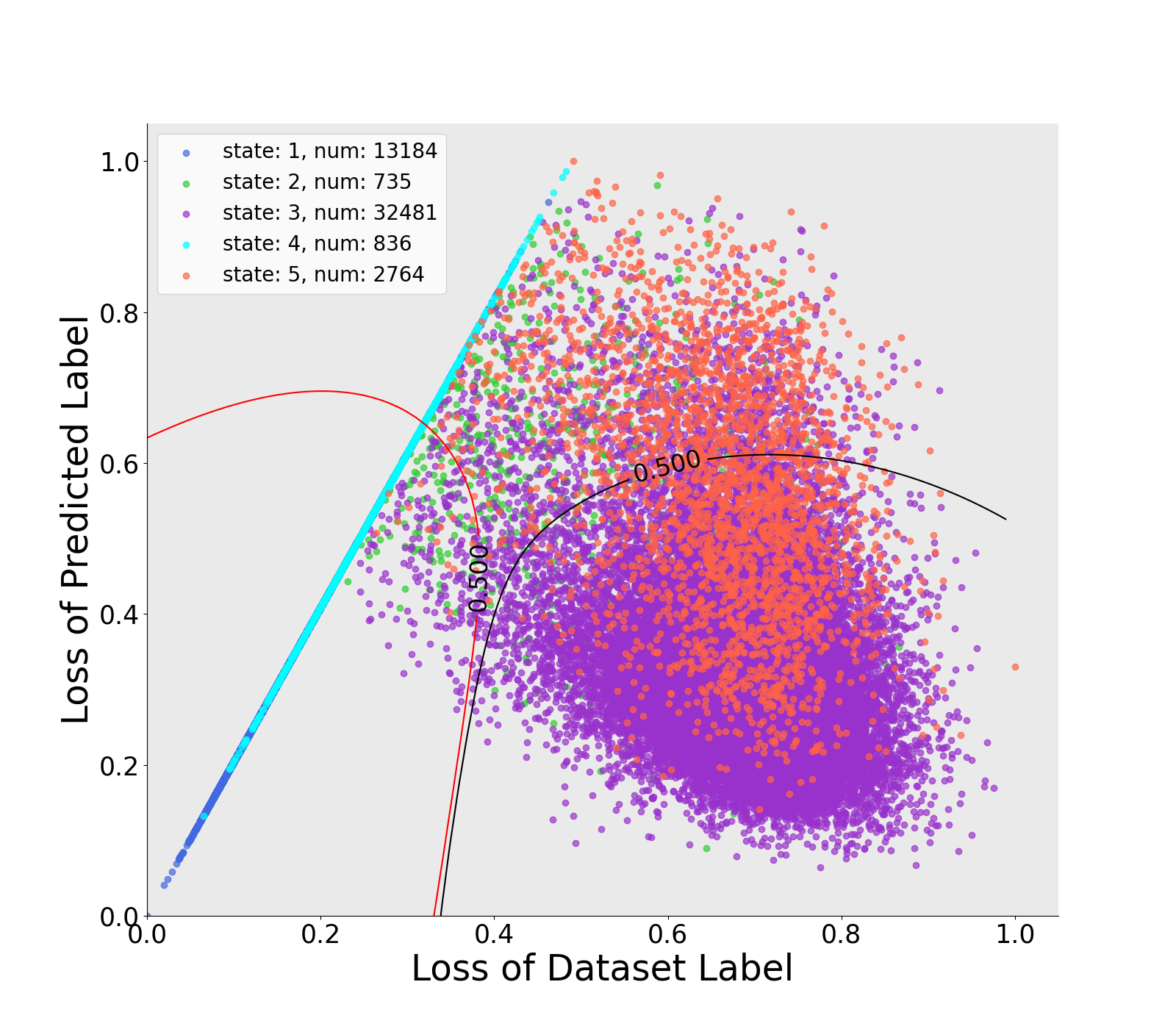}
    }

    \vspace{-10pt}
    \setcounter{subfigure}{0}
    \subfigure[Epoch 15: Cross Entropy]{
            \centering
            \includegraphics[width=4cm,height=3.5cm]{cr1/Lab-info/epoch_15/net2_losses.png}
    }
    \subfigure[Epoch 50: DST]{
            \centering
            \includegraphics[width=4cm,height=3.5cm]{cr1/Lab-info/epoch_50/net2_losses.png}
    }
    \subfigure[Epoch 100: DST]{
            \centering
            \includegraphics[width=4cm,height=3.5cm]{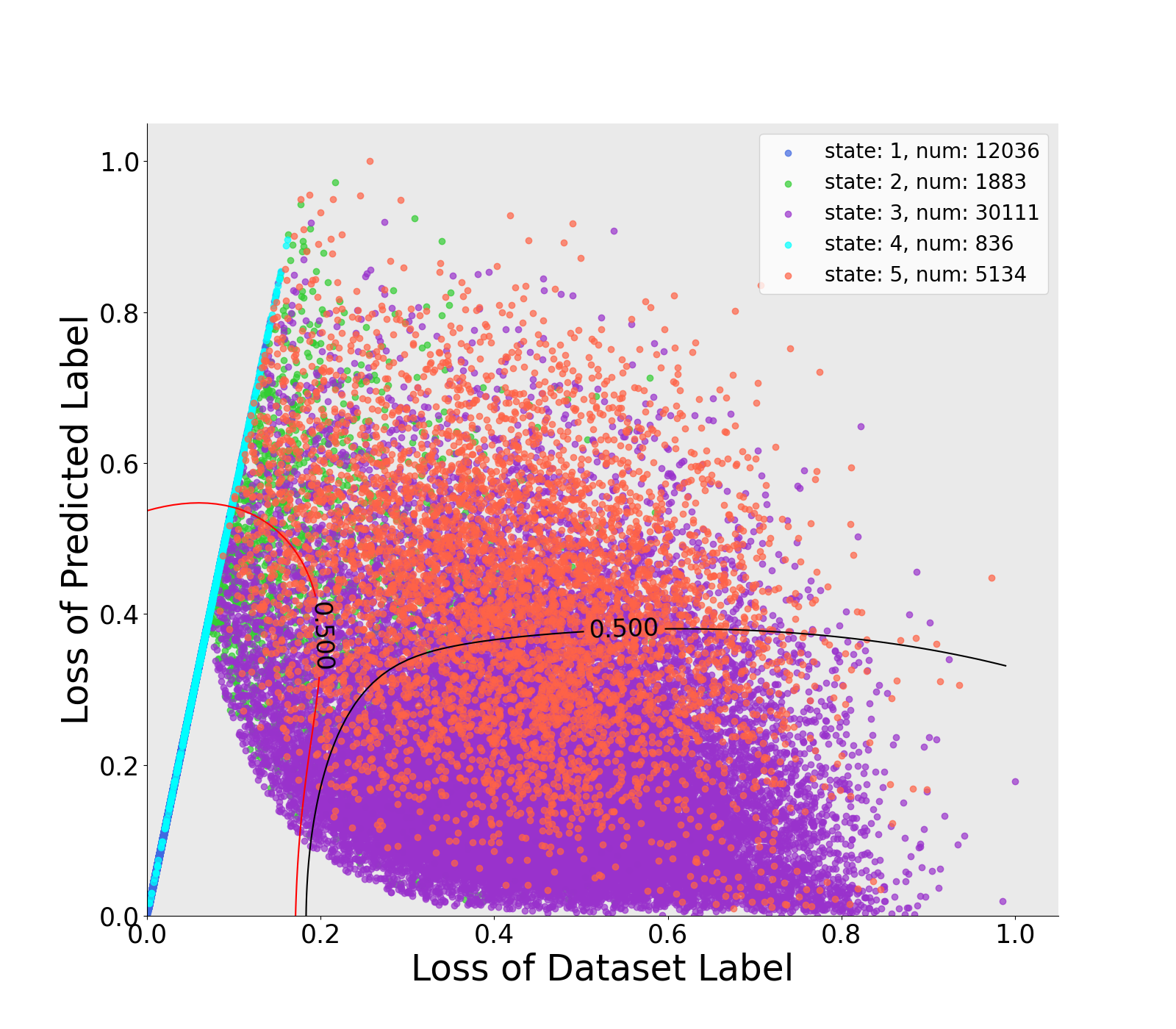}
    }
    \subfigure[Epoch 200: DST]{
            \centering
            \includegraphics[width=4cm,height=3.5cm]{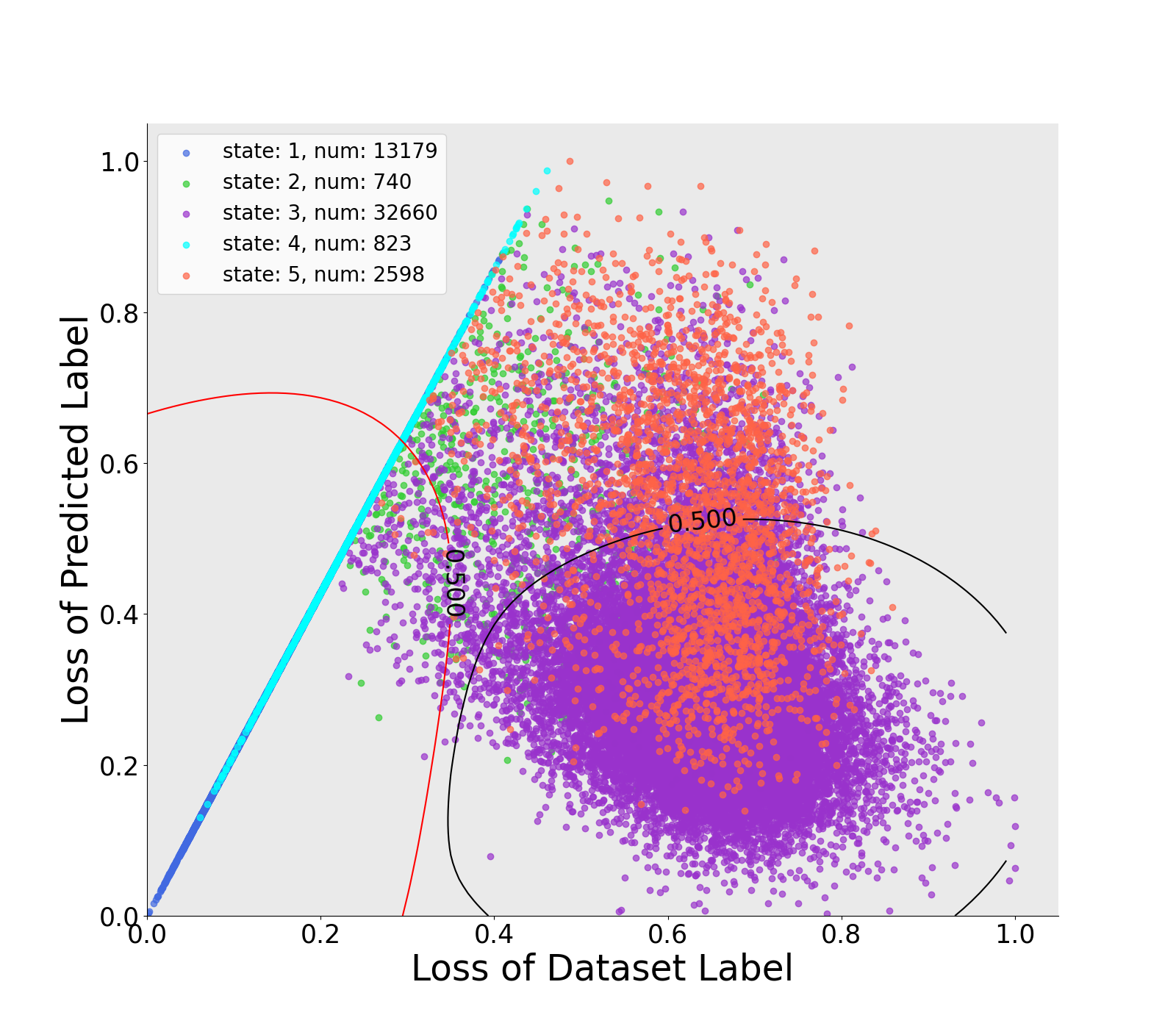}
    }
    \centering
    \caption{
        Distributions of the normalized loss on CIFAR-10 with 80\% noise ratio. Top: network A; bottom: network B.
        Dots of different colors represent different states of samples on the current epoch.
    }
    \vspace*{-15pt}
    \label{fig:distribution}
\end{figure*}

\begin{figure*}[!t]
    \centering
    \subfigure{
            \centering
            \includegraphics[width=4cm,height=3.5cm]{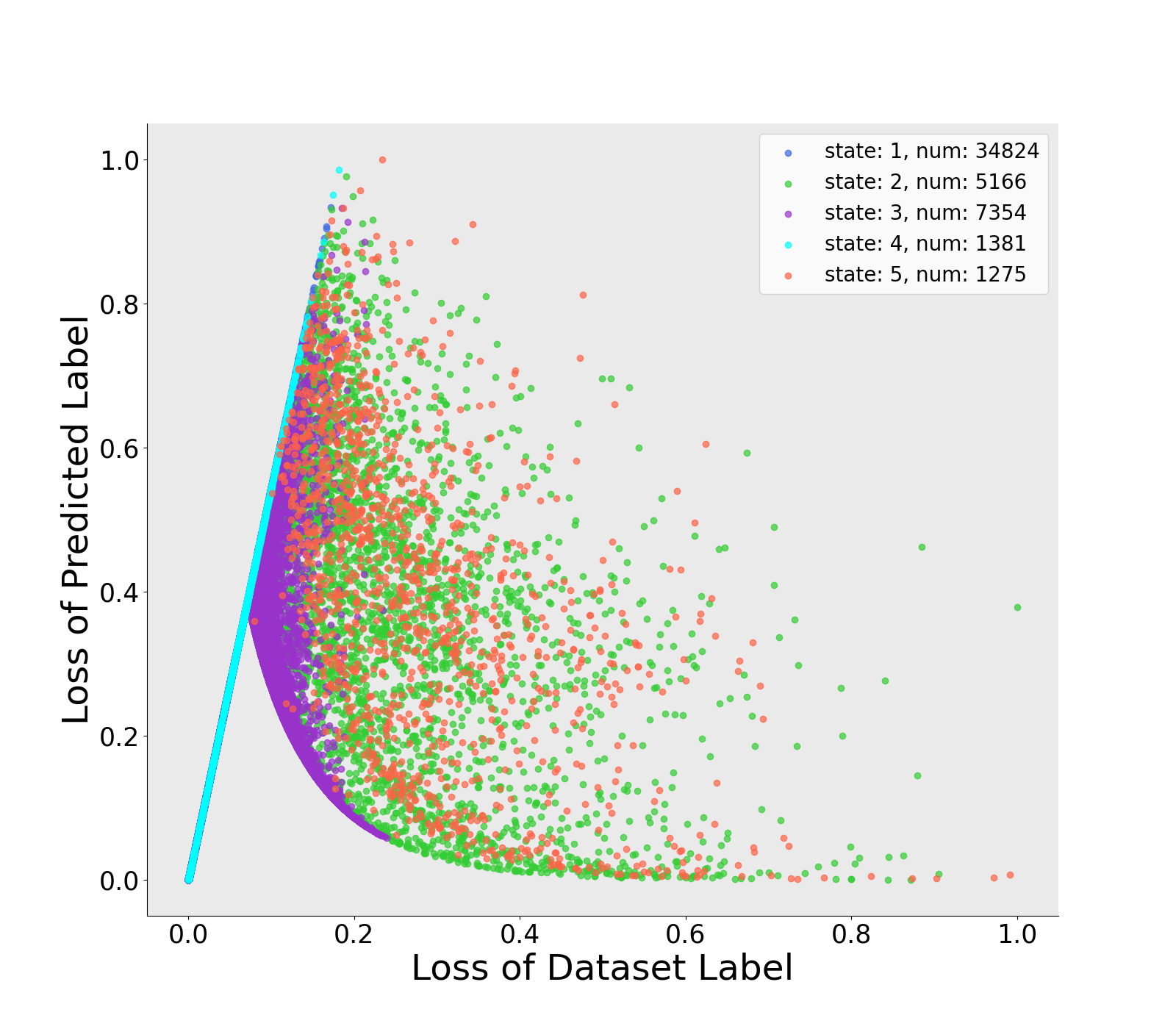}
    }
    \subfigure{
            \centering
            \includegraphics[width=4cm,height=3.5cm]{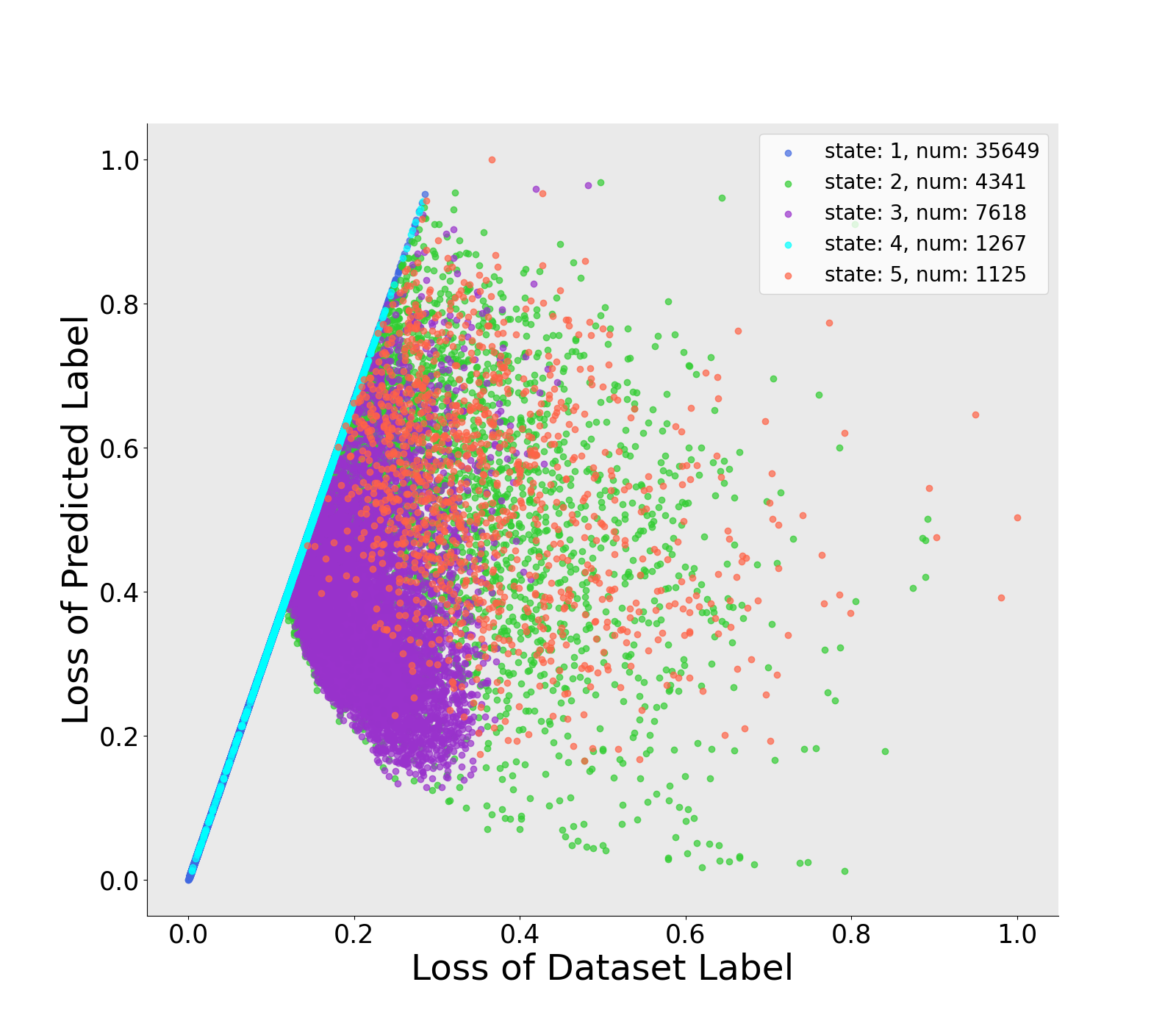}
    }
    \subfigure{
            \centering
            \includegraphics[width=4cm,height=3.5cm]{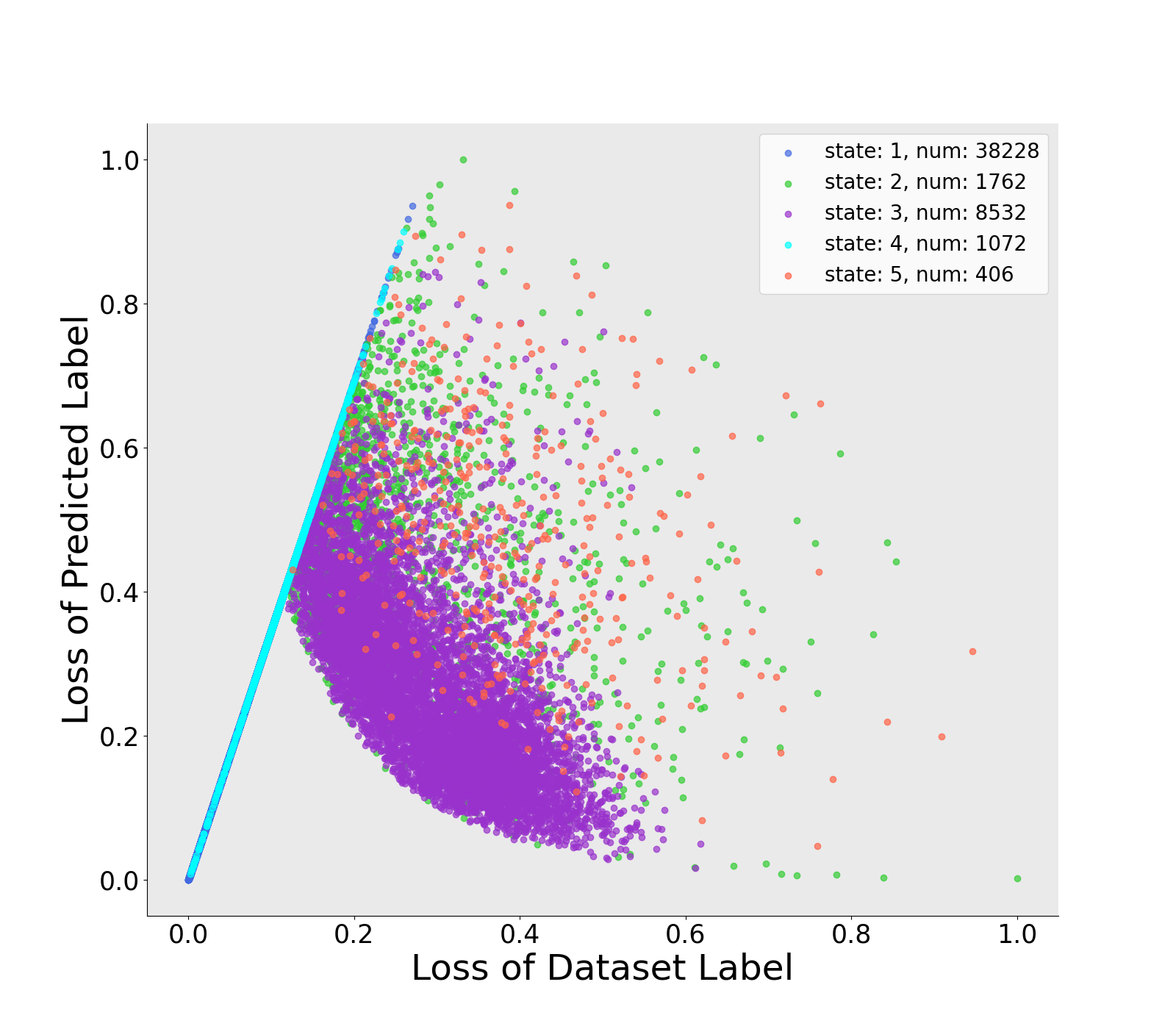}
    }
    \subfigure{
            \centering
            \includegraphics[width=4cm,height=3.5cm]{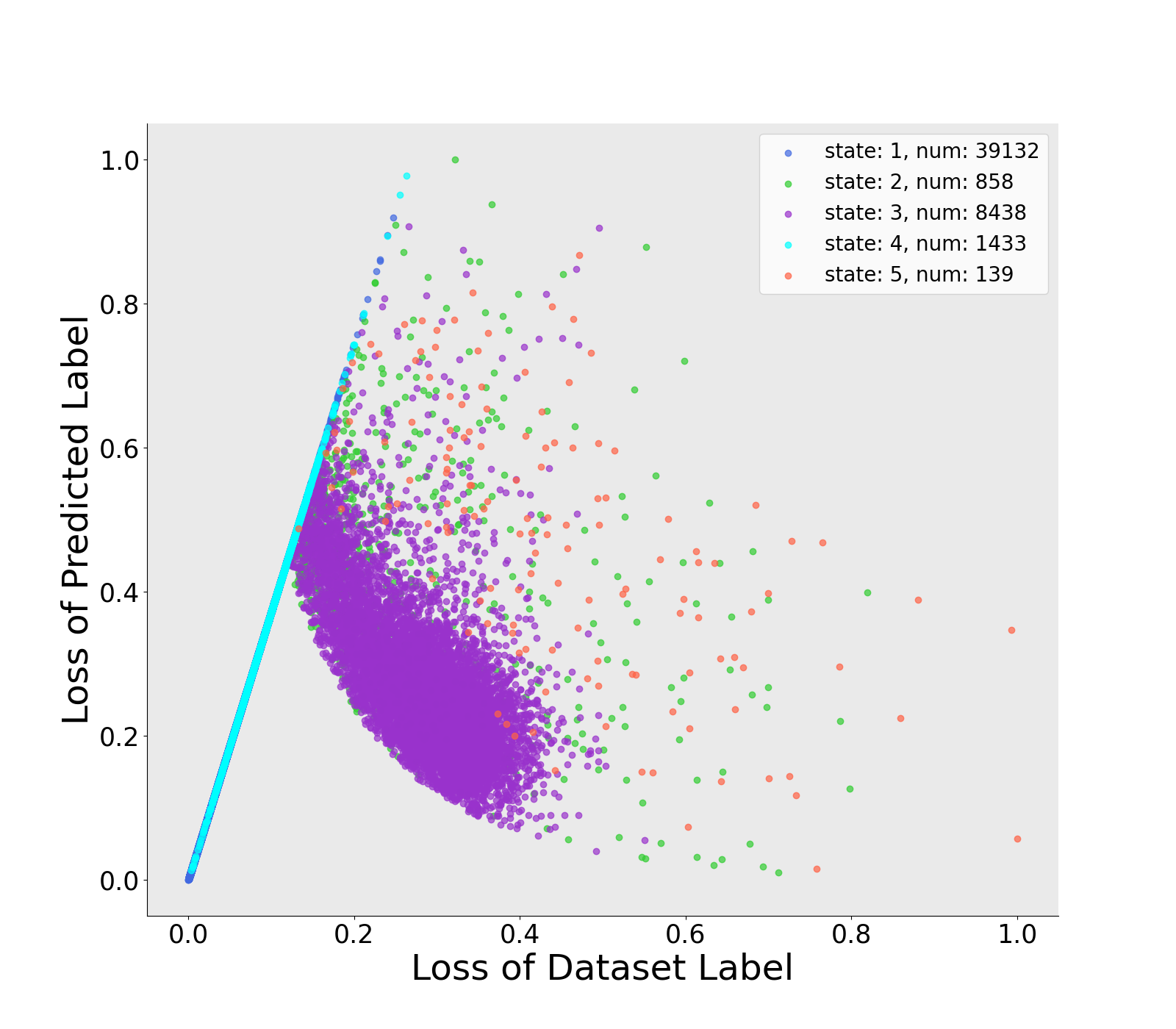}
    }

    \vspace{-10pt}
    \setcounter{subfigure}{0}
    \subfigure[Epoch 15: Cross Entropy]{
            \centering
            \includegraphics[width=4cm,height=3.5cm]{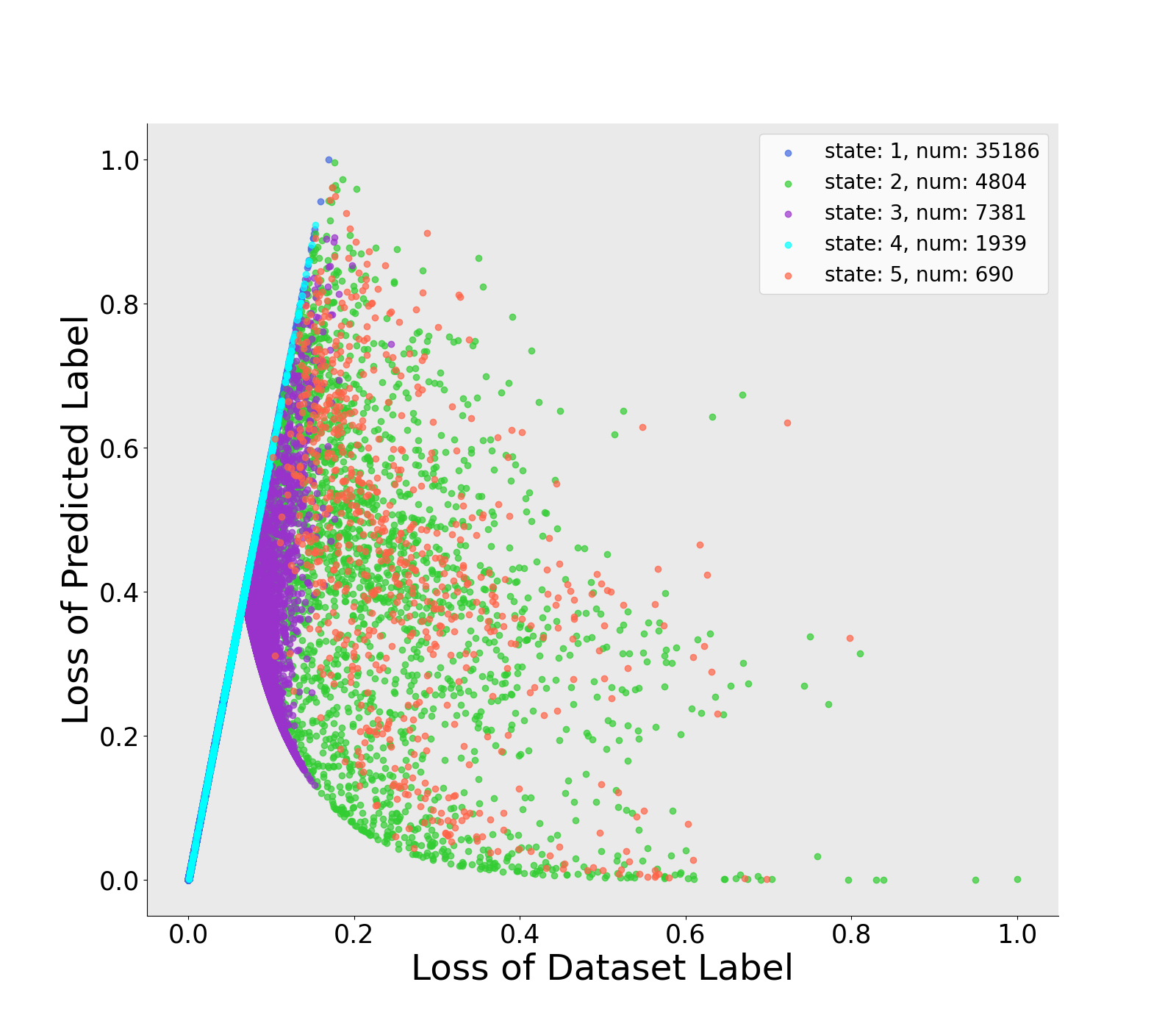}
    }
    \subfigure[Epoch 50: DST]{
            \centering
            \includegraphics[width=4cm,height=3.5cm]{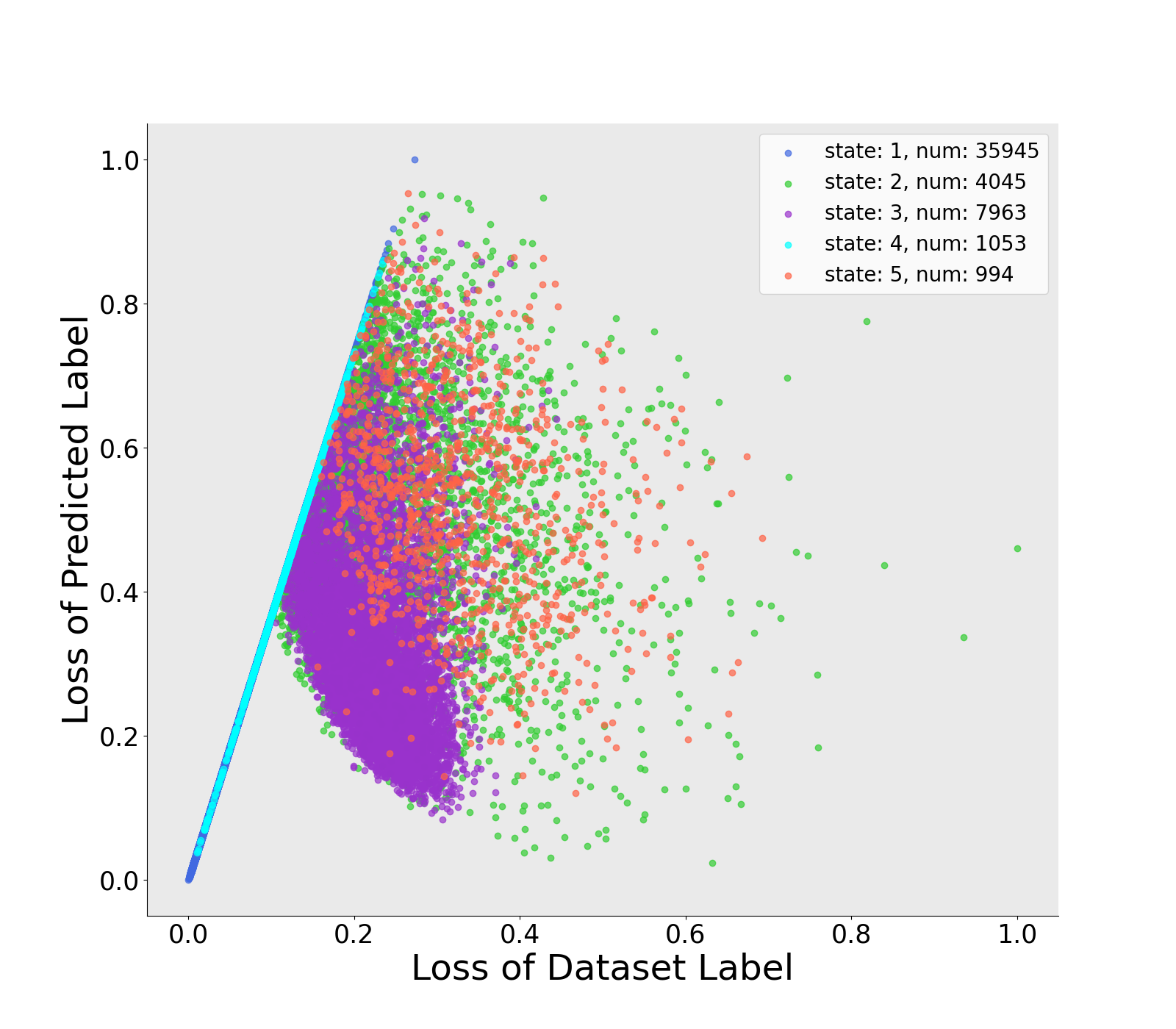}
    }
    \subfigure[Epoch 150: DST]{
            \centering
            \includegraphics[width=4cm,height=3.5cm]{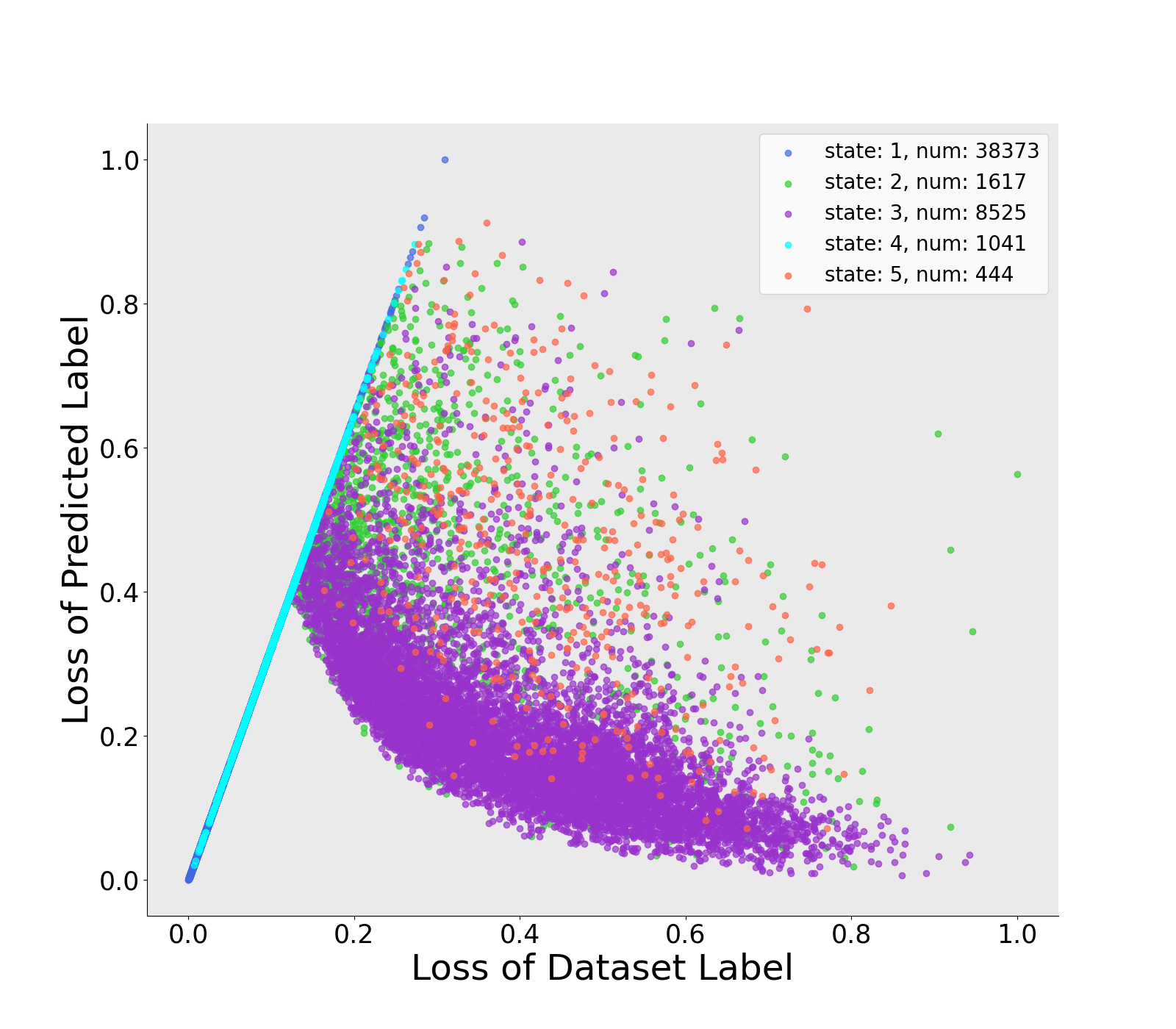}
    }
    \subfigure[Epoch 250: DST]{
            \centering
            \includegraphics[width=4cm,height=3.5cm]{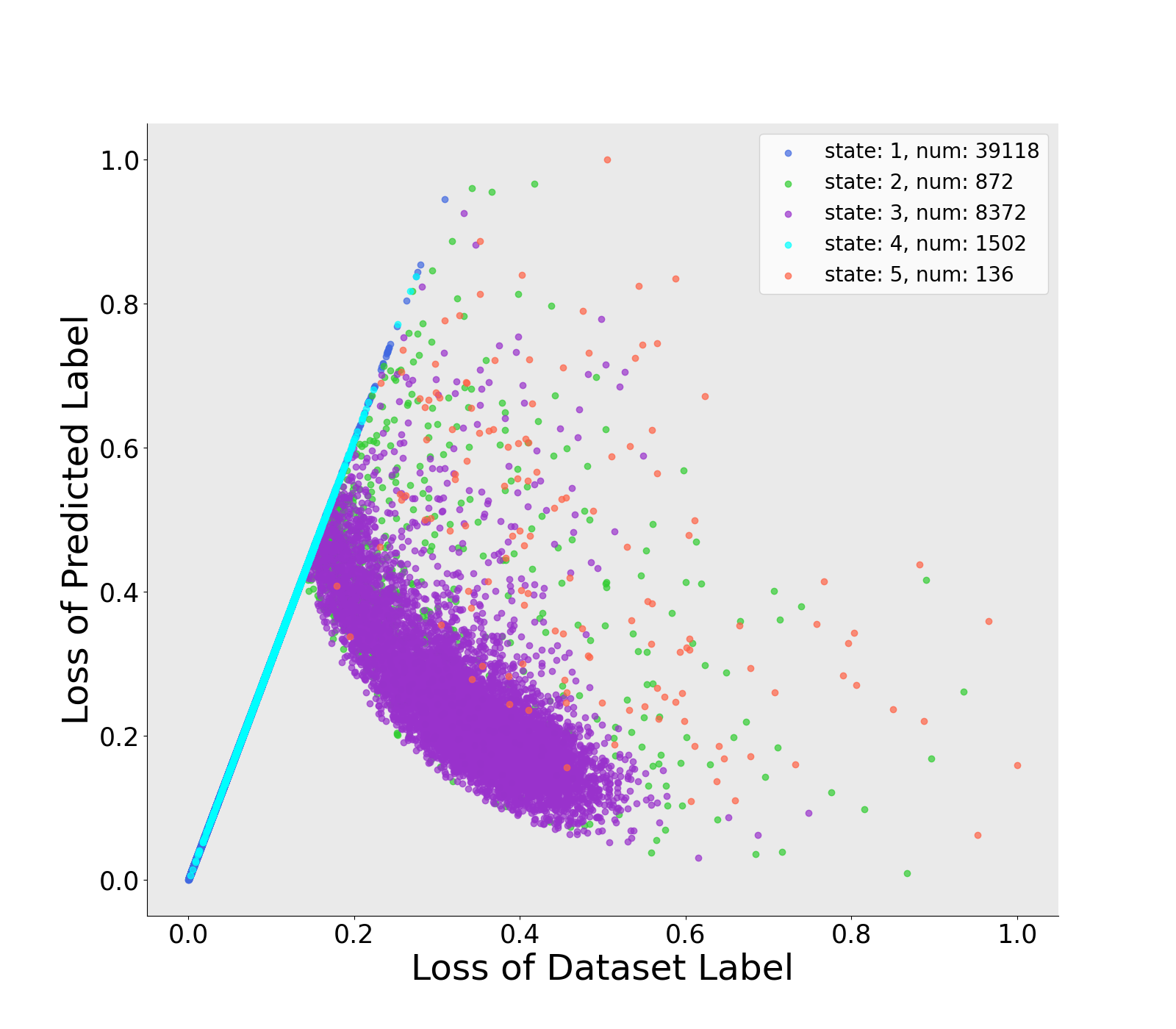}
    }
    \centering
    \caption{
        Distributions of the normalized loss on CIFAR-10 with 40\% asymmetric noise. Top: network A; bottom: network B.
        Dots of different colors represent different states of samples on the current epoch.
    }
    \vspace*{-15pt}
    \label{fig:distribution_asym}
\end{figure*}

In Figure~\ref{fig:dis_prove}, 
% 描述图像
....
In Figure~\ref{fig:distribution},
we observe that DNNs can correct predict some samples with a higher probability for the real label than other labels,
meaning that samples with state (iii) have lower $\ell_i^{prd}$ and higher $\ell_i^{nis}$ than other states.
Although samples with state (v) may have the same distribution as (iii),
the number of samples with state (iii) is much larger than (v).
Similarly, samples with state (i) have lower $\ell_i^{nis}$ than (iv).
In the case of asymmetric noise (see Figure~\ref{fig:distribution_asym}a),
The distribution of $\ell_i^{prd}$ and $\ell_i^{nis}$ is not regular and hard to be modeled by minimizing the cross entropy.
In Figure~\ref{fig:distribution_asym}b, 
the losses are shown when the model is trained with DST for 35 epoches after pre-training.
DST can distinguish the samples with state (iii) significantly and use these samples to train networks.

The Gaussian Mixture Model(GMM)~\cite{GMM} is widely used in unsupervised field due to its flexibility.
Therefore, we can fit a three components GMM to $\ell_i^{prd}$ and $\ell_i^{nis}$ by the Expectation-Maximization(EM) algorithm to distinguish correct and wrong samples.
Each sample is given a posterior probability $p(g_k|\ell_i^{nis},\ell_i^{prd})$ as $w_i^k$ by GMM, where $g_k$ is one of the Gaussian components.
We denote that $w_i^r$ is the correctly labeled probability of each sample and $w_i^{prd}$ is the correctly predicted probability of each sample.
Due to the concentrated distribution of $\ell_i^{nis}$ and $\ell_i^{prd}$,
we set the initial means for the Gaussian components.
To prevent overfitting, the convergence threshold of EM algorithm depends on the size of the dataset.
The training data is divided into a correctly labeled set, a correctly predicted set and a wrong set by setting three thresholds $\tau_k$ on $w_i^k$.
In order to address the confirmation bias~\cite{Tarvainen_NIPS_17} caused by self-learning,
we exploit co-divide~\cite{li2020dividemix} which use two networks to generate parameters and teach each other.

\subsection{Model training}

At each epoch, every sample can be relabeled by $w_i^k$ and $\tau_k$.
We train only one network at a time and fix another network which can be used to generate a new label for each sample.
Given a mini-batch of samples with their corresponding one-hot labels and probability $\mathcal{B} = \{(x_b,y_b,w_b^r,w_b^{prd});b\in(1,...,B)\}$,
We reweight the samples with $w_b^r$ and $w_b^{prd}$ and let the samples compose a new training set $\hat{\mathcal{B}}$.
Then, we exploit MixUp~\cite{mixup} augmentation which mixes two samples with the linear relationship.

To refine the label of each sample,
we linearly combine noisy label $y_b$ and prediction average value $p_b$ obtained by two networks.
Probabilities $w_b^r$ and $w_b^{prd}$ guide different samples, respectively.
Meanwhile, the wrong sample probability $w_b^u$ is produced by a uniform distribution $\mathcal{U}(0,1)$ to prevent networks from overfitting to the noisy labels:
\begin{equation}
    \tilde{y}_b =
    \begin{cases}
        w_b^r y_b+(1-w_b^r)p_b,         & \mbox{if }w_b^r \ge \tau_r         \\
        (1-w_b^{prd})y_b+w_b^{prd} p_b, & \mbox{if }w_b^{prd} \ge \tau_{prd} \\
        (1-w_b^u)y_b+w_b^u p_b,         & \mbox{others }                     \\
    \end{cases}
\end{equation}
A sharpening function used by MixMatch~\cite{mixmatch} is applied on $\tilde{y}_b$ to adjusting its temperature:
\begin{equation}
    \hat{y}_b^c=\mathrm{Sharpen}(\tilde{y}_b,T)_c=\tilde{y}_b^{c\frac{1}{T}} \bigg{/} \sum_{c=1}^C \tilde{y}_b^{c\frac{1}{T}}
\end{equation}
% MixUp
Where the $T$ is sharpening temperature.
From the sharpening function,
we acquire a new training batch $\hat{\mathcal{B}}$ in which each sample has a more refined label.
Recently MixUp~\cite{mixup} has been applied in many methods for training DNNs and achieved good result~\cite{Arazo_ICML_2019,mixmatch}.
We follow MixUp to select a pair of samples from the new batch $\hat{\mathcal{B}}$ randomly and mix them linearly:
\begin{align}
    \lambda & \sim \mathrm{Beta}(\alpha, \alpha)         \\
    \lambda & := \mathrm{max}(\lambda, 1-\lambda)        \\
    x'      & = \lambda x_1 + (1-\lambda)x_2             \\
    y'      & = \lambda \hat{y}_1 + (1-\lambda)\hat{y}_2
\end{align}
where $(x_1,x_2)$ is a pair of two samples and $(\hat{y}_1,\hat{y}_2)$ is their corresponding labels.
We utilize the cross-entropy loss on the MixUp output $\mathcal{B}'$ with $B$ samples:
\begin{equation}
    \mathcal{L}_x = - \frac{1}{B}\sum_{x,y \in \mathcal{B}'}\sum_c y_c'\log({p_{\theta}^c}(x))
\end{equation}
To prevent the networks from assigning all samples to a single class,
we apply the regularization term used in~\cite{Arazo_ICML_2019,li2020dividemix,Tanaka_CVPR_2018},
which averages the output of all samples in mini-batch to a mean value $p_{reg}^c$ (\ie $p_{reg}^c=1/C$) to address issue:
\begin{equation}
    \mathcal{L}_{reg} = \sum_c p_{reg}^c \log \bigg(p_{reg}^c\bigg/ \frac{1}{B} \sum_{x \in\mathcal{B}'} p_{\theta}^c(x)\bigg)
\end{equation}
Then, we get total loss:
\begin{equation}
    \mathcal{L}=\mathcal{L}_x + \mathcal{L}_{reg}
\end{equation}

Subsection~\ref{exp:comparison_1} compares our approach with the state-of-the-art and presents the results of experiments.

\section{Experiments}
\label{sec:experiment}

In this section, we first introduce our experiment details, 
and then compare DST with some state-of-the-arts approaches.
We also analyze the impact of MixUp and training with two networks by ablation study.

\subsection{Datasets and implementation details}
% dataset
To validate our approach, we use the following benchmark datasets, 
namely CIFAR-10, CIFAR-100~\cite{cifar} and Clothing1M~\cite{Tong_CVPR_2015}.
CIFAR-10 with 10 classes and CIFAR-100 with 100 classes contain 50K training images and 10K test images with resolution $32\times32$.
We follow previous works~\cite{MLNT,Tanaka_CVPR_2018} to generate noise labels with two types: symmetric and asymmetric.
Symmetric noise utilizes the random labels to replace the true sample labels for a percentage.
Note that there is another label noise criterion~\cite{Jiang_ICML_2018,Wang_CVPR_2018} in which the true labels is not maintained.
In Subsection~\ref{exp:comparison_1}, 
we show the results of both symmetric noise criterions with different levels of noise ranging from $20\%$ to $80\%$.
Asymmetric noise replaces the similar classes with sample labels, which mimic the label noise in real world.
We use $40\%$ because more than $50\%$ of the asymmetric noise can cause some classes become theoretically indistinguishable.

% network - cifar
We use an 18-layer PreAct Resnet~\cite{He_ECCV_2016}.
The networks are trained for 300 epochs with a batch size of 128 by SGD with a momentum of 0.9 and a weight decay of 0.0005.
We set the learning rate as 0.02 and reduce it by a factor of 5 per 100 epoch.
Before training our method, we use 15 epochs for CIFAR-10 and 30 epochs for CIFAR-100 to pretrain the networks.
We set the GMM initial kernel as $(0,0)$,$(0.5,0.5)$,$(1,0)$ and the convergence threshold of EM algorithm as $tol=20$.
The other hyperparameters across all CIFAR experiment are the same $T=0.5$, $\alpha=4$ and $\tau=0.5$.

% network - clothing1M
Clothing1M is a real-world dataset with noisy labels, 
which consists of 1 million training images from online shopping websites with noisy labels.
We use ResNet-50 with ImageNet pretrained weights to train it for 80 epochs.
Learning rate is set as 0.02 and reduced by a factor of 2 per 20 epoch.

\subsection{Comparison with the state-of-the-arts}
\label{exp:comparison_1}

\begin{table*}[!t]
    \begin{center}
        \begin{tabular*}{0.8\textwidth}{@{\extracolsep{\fill}}llccc|ccc}
            \toprule
            Dataset                                             &      & \multicolumn{3}{c|}{CIFAR-10} & \multicolumn{3}{c}{CIFAR-100}                                                                 \\
            \midrule
            Method / ratio                                      &      & 20\%                          & 50\%                          & 80\%          & 20\%          & 50\%          & 80\%          \\
            \midrule
            \multirow{2}{*}{Cross-Entropy}                      & Best & 86.8                          & 79.5                          & 62.7          & 62.1          & 46.7          & 19.8          \\
                                                                & Last & 82.5                          & 57.4                          & 26.3          & 61.0          & 36.9          & 9.0           \\
            \midrule
            \multirow{2}{*}{Co-teaching$+$\cite{Yu_ICML_2019}}  & Best & 89.5                          & 85.7                          & 67.4          & 65.6          & 51.8          & 27.9          \\
                                                                & Last & 88.2                          & 84.1                          & 45.5          & 64.1          & 45.3          & 15.5          \\
            \midrule
            \multirow{2}{*}{P-correction\cite{Yi_2019_CVPR}}    & Best & 92.4                          & 89.1                          & 77.5          & 69.4          & 57.5          & 31.1          \\
                                                                & Last & 92.0                          & 88.7                          & 76.5          & 68.1          & 56.4          & 20.7          \\
            \midrule
            \multirow{2}{*}{Meta-Learning\cite{MLNT}}           & Best & 92.9                          & 89.3                          & 77.4          & 68.5          & 59.2          & 42.4          \\
                                                                & Last & 92.0                          & 88.8                          & 76.1          & 67.7          & 58.0          & 40.1          \\
            \midrule
            \multirow{2}{*}{M-correction\cite{Arazo_ICML_2019}} & Best & 94.0                          & 92.0                          & 86.8          & 73.9          & 66.1          & 48.2          \\
                                                                & Last & 93.8                          & 91.9                          & 86.6          & 73.4          & 65.4          & 47.6          \\
            \midrule
            \multirow{2}{*}{DivideMix\cite{li2020dividemix}}    & Best & 96.1                          & 94.6                          & \textbf{93.2} & 77.3          & \textbf{74.6} & \textbf{60.2} \\
                                                                & Last & 95.7                          & 94.4                          & \textbf{92.9} & 76.9          & \textbf{74.2} & \textbf{59.6} \\
            \midrule
            \multirow{2}{*}{DST}                                & Best & \textbf{96.1}                 & \textbf{95.2}                 & 92.9          & \textbf{78.0} & 74.3          & 57.8          \\
                                                                & Last & \textbf{95.9}                 & \textbf{94.7}                 & 92.6          & \textbf{77.4} & 73.6          & 55.3          \\
            \bottomrule
        \end{tabular*}
    \end{center}
    \caption
    {
        Comparison with baselines in terms of test accuracy (\%) on CIFAR-10 and CIFAR-100 with Criterion 1.
    }
    \vspace*{-6pt}
    \label{tbl:cifar_1}
\end{table*}

We introduce some state-of-the-art methods as follow:
(1). Co-teaching$+$~\cite{co-teaching} uses the small loss of samples to train two networks each other;
(2). P-correction~\cite{Yi_2019_CVPR} optimizes the sample labels as the network parameters;
(3). Meta-Learning~\cite{MLNT} attempts to find the parameters which make model more noise-tolerant by the gradient;
(4). M-correction~\cite{Arazo_ICML_2019} uses BMM to select clean samples and improves MixUp for label noise training;
(5). DivideMix~\cite{li2020dividemix} models sample loss with GMM and improves MixMatch to achieve excellent performance in label noise training.
For the symmetric noise criterion with correct labels (Criterion 1), 
we compare our method with these baselines using the same network architecture.
For another symmetric noise criterion (Criterion 2), 
the other state-of-the-arts methods with different network architecture are compared with the DST.

% cifar : Criterion 1
\begin{table}[ht]
    \begin{center}
        \begin{tabular}{lcc}
            \toprule
            Method                             & Best          & Last          \\
            \midrule
            Cross-Entropy                      & 85.3          & 71.7          \\
            P-correction\cite{Yi_2019_CVPR}    & 88.5          & 88.1          \\
            Meta-Learning\cite{MLNT}           & 89.2          & 88.6          \\
            M-correction\cite{Arazo_ICML_2019} & 87.4          & 86.3          \\
            DivideMix\cite{li2020dividemix}    & 93.4          & 92.1          \\
            \midrule
            DST                                & \textbf{94.3} & \textbf{92.3} \\
            \bottomrule
        \end{tabular}
    \end{center}
    \caption
    {
        Comparison with baselines in terms of test accuracy (\%) on CIFAR-10 with 40\% asymmetric noise.
        We implement our method under the DivideMix setting.
    }
    \vspace*{-15pt}
    \label{tbl:cifar_asym}
\end{table}

% Compare
Table~\ref{tbl:cifar_1} shows the results on CIFAR-10 and CIFAR-100 in Criterion 1,
and the results on CIFAR-10 with 40\% asymmetric noise is shown in Table~\ref{tbl:cifar_asym}.
In the tables, 
"Best" is the best test accuracy across training process and "Last" is the average accuracy of the last 10 epochs.
As we can see, the performance of DST is the same or superior to the other baselines. 
DST works a little worse than DivideMix on CIFAR-100 with 80\% noise ratio, 
but DST achieves comparable performance with DivideMix across the other noise ratios.
DST works much better than all other state-of-the-art methods on CIFAR-10 with 40\% asymmetric noise ratio.
Table~\ref{tbl:cifar_2} shows the comparison in Criterion 2.
DST outperforms baselines by a large margin on CIFAR-10 with 80\% noise ratio. 
However, DST has the same problems as Criterion 1 on CIFAR-100 with 80\% noise ratio, 
and our explanation is given in the next paragraph.

\begin{table}[!t]
    \centering
    \begin{minipage}[!t]{\columnwidth}
        \centering
        \begin{tabular}{lcccc}
            \toprule
            Method / ratio                  & 20\%          & 40\%          & 60\%          & 80\%          \\
            \midrule
            MentorNet~\cite{Jiang_ICML_2018}    & 92.0          & 89.0          & -             & 49.0          \\
            D2L~\cite{Ma_ICML_2018}             & 85.1          & 83.4          & 72.8          & -             \\
            Reweight~\cite{Ren_ICML_2018}       & 86.9          & -             & -             & -             \\
            Abstention~\cite{Abstention}        & 93.4          & 90.9          & 87.6          & 70.8          \\
            M-correction~\cite{Arazo_ICML_2019} & 94.0          & 92.8          & 90.3          & 74.1          \\
            DivideMix~\cite{li2020dividemix}    & \textbf{96.2} & 94.9          & \textbf{94.3} & 79.8          \\
            \midrule
            DST                                 & 96.0          & \textbf{95.3} & 94.1          & \textbf{87.5} \\
            \bottomrule
        \end{tabular}
    \end{minipage}
    \\ [6pt]
    \begin{minipage}[!t]{\columnwidth}
        \centering
        \begin{tabular}{l cccc}
            \toprule
            Method / ratio                  & 20\%          & 40\%          & 60\%          & 80\%          \\
            \midrule
            MentorNet~\cite{Jiang_ICML_2018}    & 73.0          & 68.0          & -             & 35.0          \\
            D2L~\cite{Ma_ICML_2018}             & 62.2          & 52.0          & 42.3          & -             \\
            Reweight~\cite{Ren_ICML_2018}       & 61.3          & -             & -             & -             \\
            Abstention~\cite{Abstention}        & 75.8          & 68.2          & 59.4          & 34.1          \\
            M-correction~\cite{Arazo_ICML_2019} & 73.7          & 70.1          & 59.5          & 45.5          \\
            DivideMix~\cite{li2020dividemix}    & 77.2          & 75.2          & \textbf{72.0} & \textbf{60.0} \\
            \midrule
            DST                                 & \textbf{77.5} & \textbf{75.3} & 70.4          & 57.7          \\
            \bottomrule
        \end{tabular}
    \end{minipage}
    \\ [12pt]
    \caption
    {
        Comparison with baselines in terms of test accuracy (\%) with Criterion 2. 
        Top : CIFAR-10, 
        bottom : CIFAR-100.
    }
    \vspace*{-15pt}
    \label{tbl:cifar_2}
\end{table}

In the experiments of CIFAR, due to the existence of noisy labels in the dataset, 
it is difficult to improves the performance of methods which have achieved very good results.
Unfortunately, in the case of high noise ratios,
our method can produce a small amount of mistakes leading to the confirmation bias problem, 
especially in the dataset with more classes like CIFAR-100,
in which we have a small gap with the state-of-the-art method~\cite{li2020dividemix}.
However, DST still outperforms the other methods across all noise ratios.

\begin{figure*}[!t]
    \centering
    \subfigure{
        \centering
        \includegraphics[width=12cm]{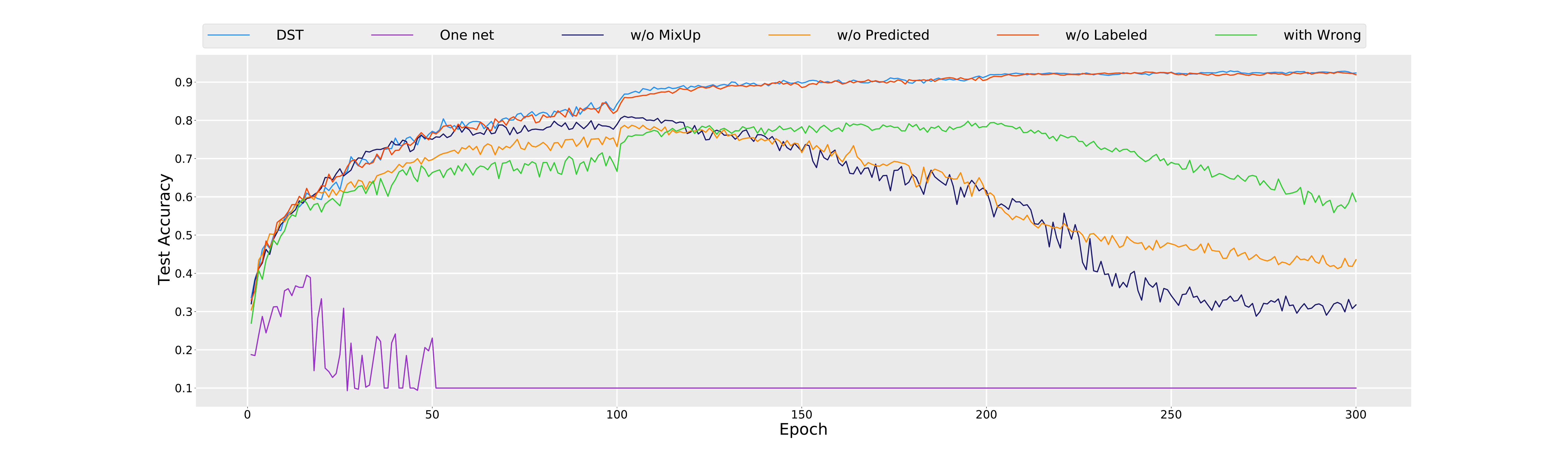}
    }
    \vspace{-11pt}

    \subfigure[20\% noise ratio]{
            \centering
            \includegraphics[width=4cm,height=3.5cm]{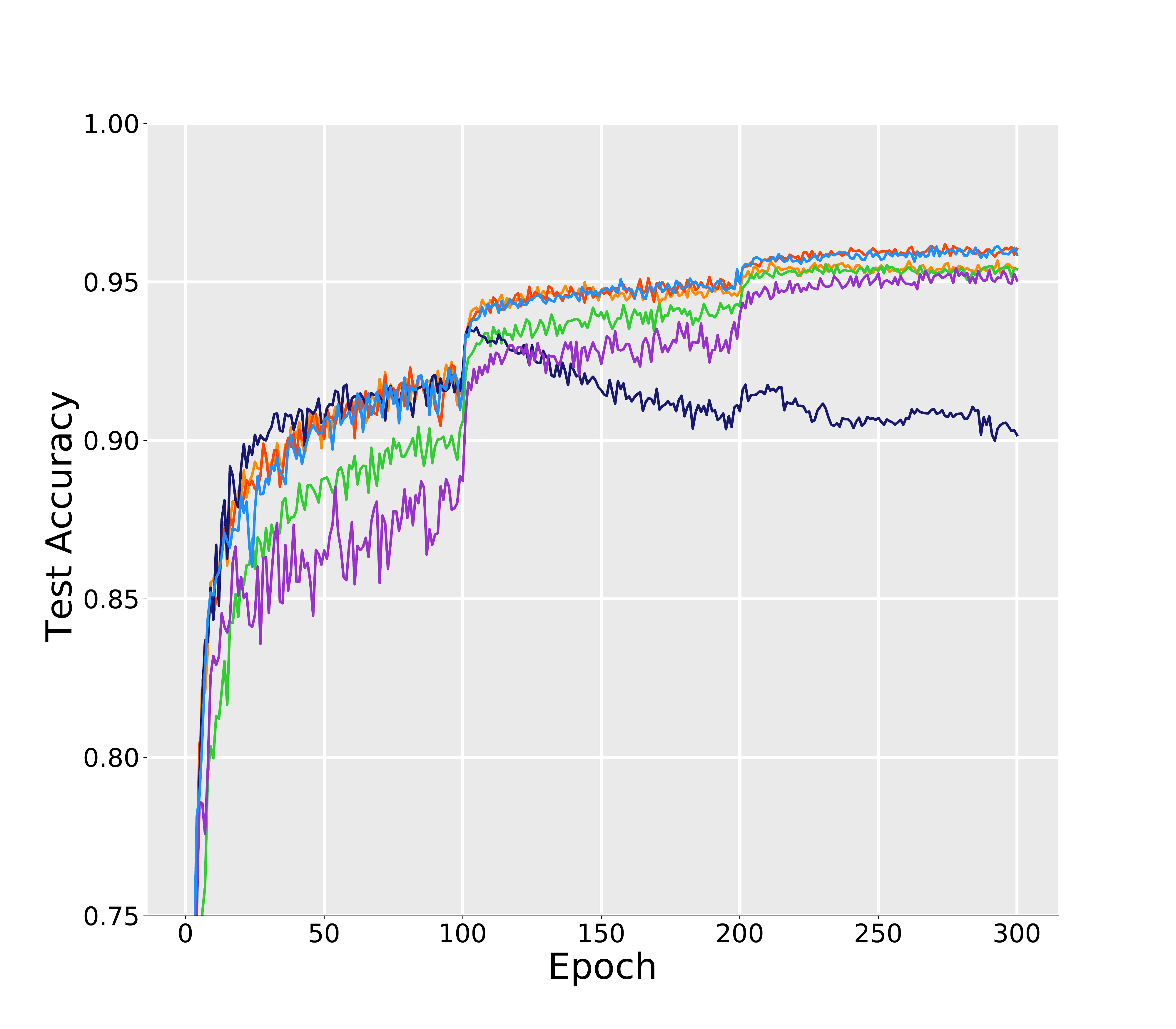}
    }
    \subfigure[50\% noise ratio]{
            \centering
            \includegraphics[width=4cm,height=3.5cm]{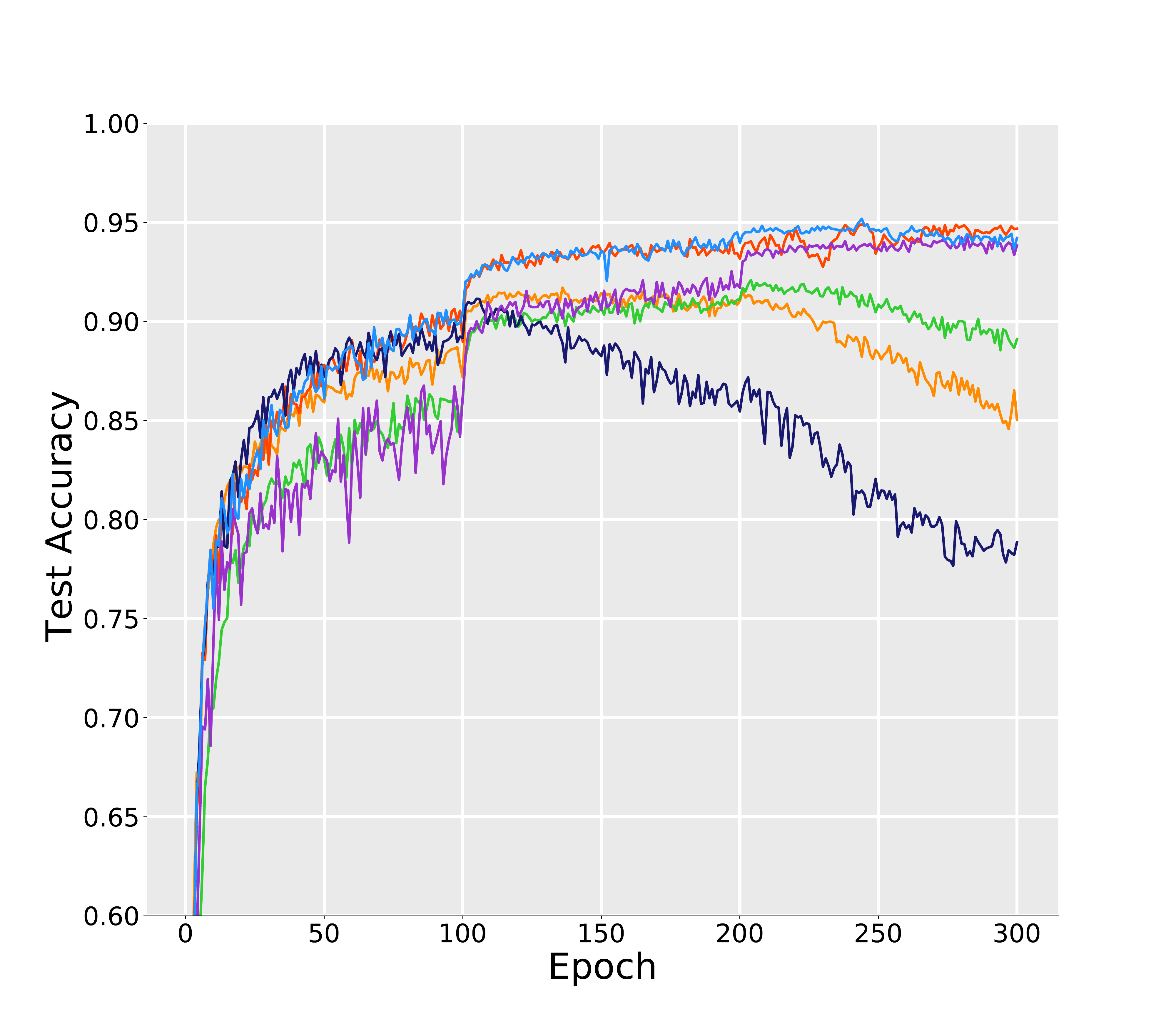}
    }
    \subfigure[80\% noise ratio]{
            \centering
            \includegraphics[width=4cm,height=3.5cm]{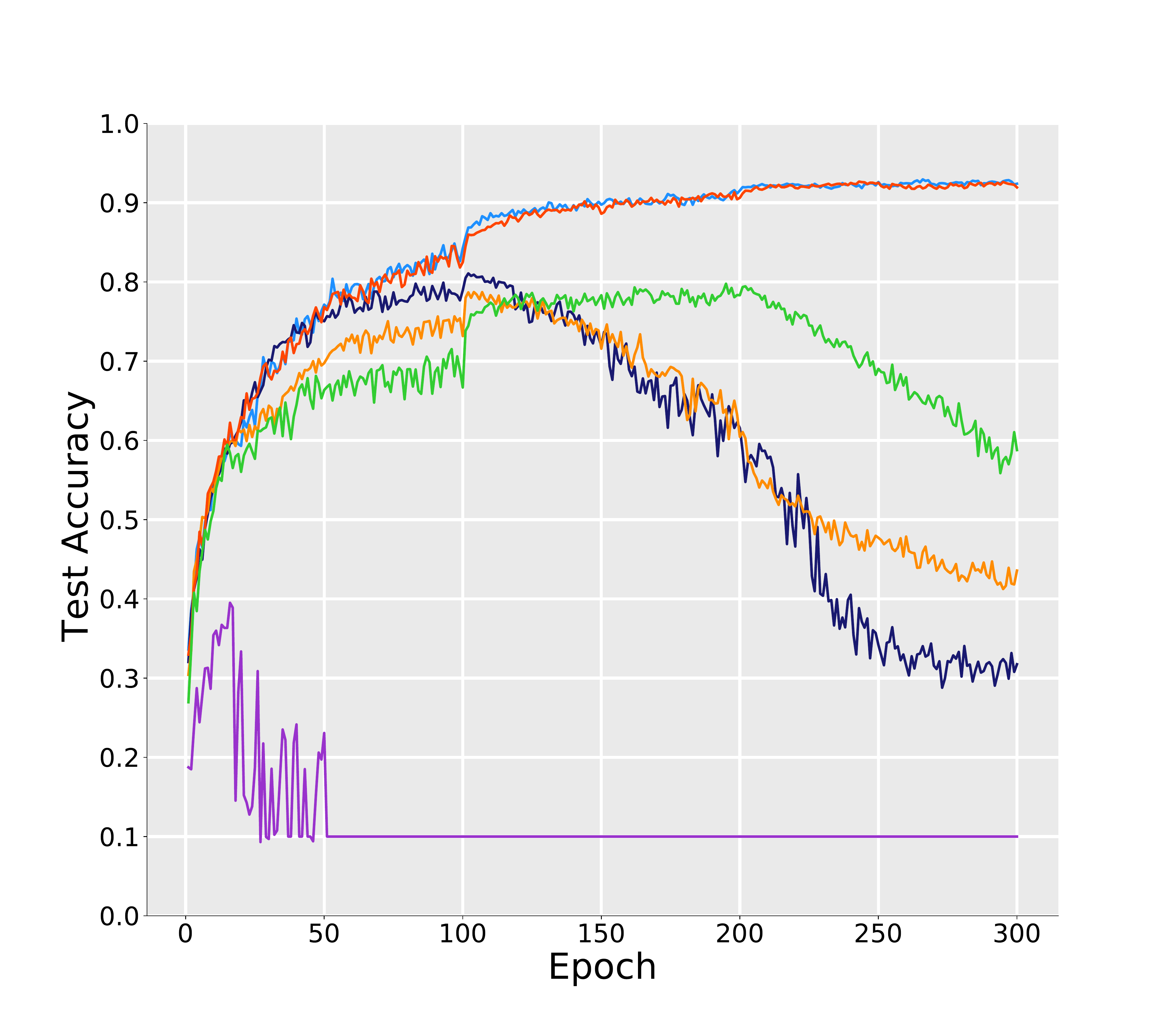}
    }

    \centering
    \caption{
        Ablation study results on CIFAR-10. Test accuracy(\%) vs. epochs.
    }
    \vspace*{-6pt}
    \label{fig:cifar_10_as}
\end{figure*}

\begin{table}[!t]
    \begin{center}
        \begin{tabular}{lc}
            \toprule
            Method                                & Test Accuracy  \\
            \midrule
            Cross-Entropy                         & 69.32          \\
            F-correction~\cite{Giorgio_CVPR_2017} & 69.84          \\
            M-correction~\cite{Arazo_ICML_2019}   & 71.00          \\
            Joint-Optim~\cite{Tanaka_CVPR_2018}   & 72.23          \\
            Meta-Learning~\cite{MLNT}             & 73.47          \\
            P-correction~\cite{Yi_2019_CVPR}      & 73.49          \\
            DivideMix~\cite{li2020dividemix}      & \textbf{74.76} \\
            \midrule
            DST                                   & 73.67          \\
            \bottomrule
        \end{tabular}
    \end{center}
    \caption
    {
        Comparison with state-of-the-art methods in test accuracy (\%) on Clothing1M.
    }
    \vspace*{-15pt}
    \label{tbl:clothing1m}
\end{table}

Table~\ref{tbl:clothing1m} shows the results on Clothing1M dataset. 
DST obtains 73.67\% test accuracy, which is lower than the state-of-the-art~\cite{li2020dividemix}.
Using a pre-trained network and a small learning rate can easily fits label noise~\cite{Arazo_ICML_2019}.
Therefore, we believe that the data partition ability of DST is well limited.
However, compared with Cross-Entropy, DST has a good improvement.

\subsection{Ablation study}

To understand what makes DST successful, we attempt to remove the MixUp, 
use one network for self-learning and use different modules of the training (\ie different partition set) to train networks.
We set up the experiments on CIFAR-10 in Criterion 1 with noise ratios (20\%, 50\% and 80\%). 
The results of the experiments is shown in Table~\ref{tbl:cifar_as}.

\begin{table}[!t]
    \begin{center}
        \begin{tabular}{llccc}
            \toprule
            Method / ratio                           &      & 20\%          & 50\%          & 80\%          \\
            \midrule
            \multirow{2}{*}{DST}                     & Best & \textbf{96.1} & \textbf{95.2} & \textbf{92.9} \\
                                                     & Last & \textbf{95.9} & 94.1          & \textbf{92.6} \\
            \midrule
            \multirow{2}{*}{DST w/o MixUp}           & Best & 93.5          & 91.1          & 81.1          \\
                                                     & Last & 90.3          & 78.7          & 31.3          \\
            \midrule
            \multirow{2}{*}{DST with $\theta^{(1)}$} & Best & 95.4          & 94.3          & 10.0          \\
                                                     & Last & 95.1          & 93.8          & 10.0          \\
            \midrule
            DST w/o                                  & Best & 95.7          & 91.7          & 78.7          \\
            the correctly predicted set              & Last & 95.4          & 85.4          & 42.5          \\
            \midrule
            DST w/o                                  & Best & 96.1          & 94.9          & 92.6          \\
            the correctly labeled set                & Last & 95.9          & \textbf{94.7} & 92.3          \\
            \midrule
            DST with                                 & Best & 95.5          & 92.1          & 79.7          \\
            the wrong set                            & Last & 95.4          & 89.1          & 58.1          \\
            \bottomrule
        \end{tabular}
    \end{center}
    \caption
    {
        Ablation study results in terms of test accuracy (\%) on CIFAR-10 in Criterion 1.
        First, DST is trained without MixUp.
        Second, DST uses one network to teach itself.
        Third, DST is trained with the correctly predicted set as the wrong set.
        Fourth, DST is trained with the correctly labeled set as the wrong set, which has best accuracy at "Last" in 50\% noise ratio.
        Last, DST is trained with all sets as the wrong set.
    }
    \vspace*{-15pt}
    \label{tbl:cifar_as}
\end{table}

From the results, we clarify some details in the ablation study. 
First, for DST, two networks are obviously superior to one network.
In a high noise ratio, one network can not be trained because of the confirmation bias problem, 
but two networks address this problem.
Second, MixUp is also a good approach to address the confirmation bias problem.
In Figure~\ref{fig:cifar_10_as}, we get the lower accuracy of "Last" when training DST wihtout MixUp across all noise ratios.
Finally, Date Selection is very important to DST, especially the module of the correctly predicted training.
We guess that the random factors of the wrong samples including the correctly labeled samples replaces the module of the correctly labeled training.
In the case of high noise ratio,
the correctly labeled set is a small part of the entire training set, 
which has little effect in the second half of training process.
In the low noise ratio, 
since most samples is in the state (i) introduced in~\ref{med:data_selection},
the random module of the wrong sample training is almost same as the correctly labeled training.
Generally, the correctly labeled training has a improvement on DST,
and the module of the correctly predicted training is essential for DST.

\section{Conclusion}
\label{sec:conclusion}

This paper presents DST on learning with noisy labels.
Our method fits a Gaussian mixture model to the cross-entropy loss between the sample and its two labels.
Meanwhile, we use two networks to teach each other by the probabilities from GMM. 
Through many experiments on different datasets, 
we show that DST can achieve the same or better performance compared with state-of-the-art methods.
We further propose to use semi-supervised learning to train the wrong samples generated by GMM.
In addition, we are interested in finding a new unsupervised method to improve the capability of data selection.

{\small
\bibliographystyle{ieee_fullname}
\bibliography{egbib}
}

\end{document}